\documentclass[10pt,twocolumn,letterpaper]{article}

\usepackage{iccv}
\usepackage{times}
\usepackage{epsfig}
\usepackage{graphicx}
\usepackage{amsmath}
\usepackage{amssymb}
\usepackage{booktabs}
\usepackage{multirow}
\usepackage{float}
\usepackage{subfigure}

\usepackage{xcolor}

\usepackage[pagebackref=true,breaklinks=true,letterpaper=true,colorlinks,bookmarks=false]{hyperref}

\iccvfinalcopy % *** Uncomment this line for the final submission

\def\iccvPaperID{3822} % *** Enter the ICCV Paper ID here

\ificcvfinal\fi

\begin{document}

%%%%%%%%% TITLE
\title{Pyramid R-CNN:\\ Towards Better Performance and Adaptability for 3D Object Detection}

\author{Jiageng Mao $^1$
%\and
\hspace{2mm}
Minzhe Niu $^2$
%\and
\hspace{2mm}
Haoyue Bai $^3$
%\and
\hspace{2mm}
Xiaodan Liang $^4$$^{\dag}$
%\and
\hspace{2mm}
Hang Xu $^2$
%\and
\hspace{2mm}
Chunjing Xu $^2$
}

\maketitle
% Remove page # from the first page of camera-ready.
\ificcvfinal\thispagestyle{empty}\fi

%%%%%%%%% ABSTRACT
\begin{abstract}
We present a flexible and high-performance framework, named Pyramid R-CNN, for two-stage 3D object detection from point clouds. Current approaches generally rely on the points or voxels of interest for RoI feature extraction on the second stage, but cannot effectively handle the sparsity and non-uniform distribution of those points, and this may result in failures in detecting objects that are far away. To resolve the problems, we propose a novel second-stage module, named pyramid RoI head, to adaptively learn the features from the sparse points of interest. The pyramid RoI head consists of three key components. Firstly, we propose the RoI-grid Pyramid, which mitigates the sparsity problem by extensively collecting points of interest for each RoI in a pyramid manner. Secondly, we propose RoI-grid Attention, a new operation that can encode richer information from sparse points by incorporating conventional attention-based and graph-based point operators into a unified formulation. Thirdly, we propose the Density-Aware Radius Prediction (DARP) module, which can adapt to different point density levels by dynamically adjusting the focusing range of RoIs. Combining the three components, our pyramid RoI head is robust to the sparse and imbalanced circumstances, and can be applied upon various 3D backbones to consistently boost the detection performance. Extensive experiments show that Pyramid R-CNN outperforms the state-of-the-art 3D detection models by a large margin on both the KITTI dataset and the Waymo Open dataset.
\end{abstract}

\let\thefootnote\relax\footnotetext{$^1$ The Chinese University of Hong Kong $^2$ Huawei Noah's Ark Lab \\ $^3$ HKUST $^4$ Sun Yat-Sen University \\ $^{\dag}$ Corresponding author: \url{xdliang328@gmail.com}}

%%%%%%%%% BODY TEXT
\section{Introduction}
3D object detection is a key component of perception systems for robotics and autonomous driving, aiming at detecting vehicles, pedestrians, and other objects with 3D point clouds as input. In this paper, we propose a general two-stage 3D detection framework, named Pyramid R-CNN, which can be applied on multiple 3D backbones to enhance the detection adaptability and performance.

\begin{figure}[!t]
\centering
\includegraphics[width=0.45\textwidth]{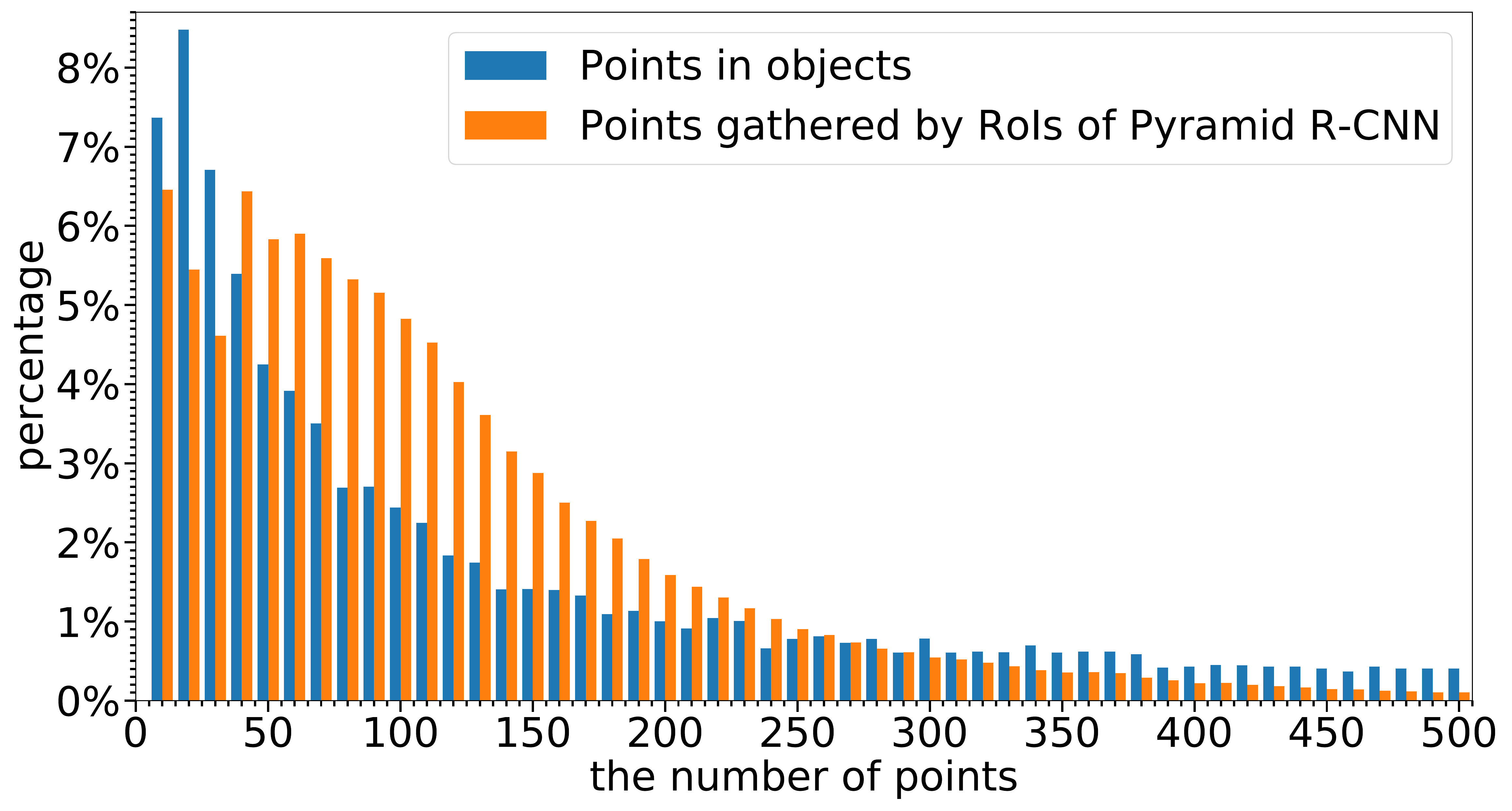}
\caption{Statistical results on the KITTI dataset. Blue bars denote the distribution of the number of object points. Orange bars denote the distribution of the number of points gathered by RoIs in Pyramid R-CNN. Our approach can mitigate the sparsity and imbalanced distribution problems of point clouds.}
\label{fig_intro}
\vspace{-6mm}
\end{figure}

Among the existing 3D detection frameworks, two-stage detection models~\cite{yang2019std, shi2020points, shi2020pv, deng2020voxel, shi2021pv} surpass most single-stage 3D detectors~\cite{zhou2018voxelnet, yan2018second, lang2019pointpillars, mao2021one, shi2019pointrcnn} with remarkable margins owing to the RoI refinement stage. Different from the 2D counterparts~\cite{girshick2014rich, girshick2015fast, ren2016faster, he2017mask, cai2018cascade} which apply RoIPool~\cite{girshick2015fast} or RoIAlign~\cite{he2017mask} to crop dense feature maps on the second stage, the 3D detection models generally perform various RoI feature extraction operations on the \textit{Points of Interest}. For example, Point R-CNN~\cite{shi2019pointrcnn} utilizes a point-based backbone to generate 3D proposals, treats the points near the proposals as Points of Interest and applies Region Pooling on those sparse points for box refinement; Part-$A^{2}$ Net~\cite{shi2020points} utilizes a voxel-based backbone for proposal generation, uses the upsampled voxel points as Points of Interest, and applies sparse convolutions on those voxel points for each RoI; PV-RCNN~\cite{shi2020pv} encodes the whole scene into a set of keypoints, and utilizes keypoints as Points of Interest for RoI-grid Pooling. Those Points of Interest originate from raw point clouds and contain rich fine-grained information, which is required for the RoI refinement stage.

However, the Points of Interest inevitably suffer from the sparsity and non-uniform distribution characteristics of input point clouds. As is demonstrated by the statistical results on the KITTI dataset~\cite{geiger2013vision} in Figure~\ref{fig_intro}: 1) Point clouds can be quite sparse in certain objects. More than $7\%$ of total objects have less than 10 points, and their visualized shapes are mostly incomplete. Thus it is hard to identify their categories without enough context information. 2) The distribution of object points is extremely imbalanced. The number of object points ranges from less than $10$ to more than $500$ on KITTI, and current RoI operations cannot handle the imbalanced conditions effectively. 3) The number of Points of Interest only accounts for a small proportion of input points or voxels, \eg $2k$ keypoints in~\cite{shi2020pv} relative to the $15k$ total input points, which exacerbates the above problems.

To overcome the above limitations, we propose Pyramid R-CNN, a general two-stage 3D detection framework that can effectively detect objects and adapt along with environmental changes. Our main contribution lies in the design of a novel RoI feature extraction head, named pyramid RoI head, which can be applied on multiple 3D backbones and Points of Interest. pyramid RoI head consists of three key components. Firstly, we propose RoI-grid Pyramid. Given the observation that Points of Interest inside RoIs are too sparse for object recognition, our RoI-grid Pyramid captures more Points of Interest outside RoIs while still maintaining fine-grained geometric details, by extending the standard one-level RoI-grid to a pyramid structure. Secondly, we propose RoI-grid Attention, an effective operation to extract RoI-grid features from Points of Interest. RoI-grid Attention leverages the advantages of the graph-based and attention-based point operators by combining those formulas into a unified formulation, and it can adapt to different sparsity situations by dynamically attending to the crucial Points of Interest near the RoIs. Thirdly, we propose the Density-Aware Radius Prediction (DARP) module, which can predict the feature extraction radius of each RoI, conditioning on the neighboring distribution of Points of Interest. Thus we can address the imbalanced distribution problem by adaptively adjusting the focusing range for each RoI. Combining all the above components, the pyramid RoI head shows adaptability to different point cloud sparsity levels and can accurately detect the 3D objects with only a few points. Our Pyramid R-CNN is compatible with the point-based~\cite{shi2019pointrcnn}, voxel-based~\cite{shi2020points} and point-voxel-based~\cite{shi2020pv} frameworks, and significantly boosts the detection accuracy.

We summarize our key contributions as follows: \\
\indent 1) We propose Pyramid R-CNN, a general two-stage framework that can be applied on multiple backbones for accurate and robust 3D object detection.\\
\indent 2) We propose the pyramid RoI head, which combines the RoI-grid Pyramid, RoI-grid Attention, and the Density-Aware Radius Prediction (DARP) module together to mitigate the sparsity and non-uniform distribution problems.\\ 
\indent 3) Pyramid R-CNN consistently outperforms the baselines, achieves $82.08\%$ moderate car mAP on the KITTI dataset, and ranks $1^{st}$ among the LiDAR-only methods on the Waymo \textit{test} leaderboard for vehicle detection.

\section{Related Work}

\noindent\textbf{Single-stage 3D Object Detection.} Single-stage methods can be divided into $3$ streams, \ie, point-based, voxel-based and pillar-based. The point-based single-stage detectors generally take the raw points as input, and apply set abstraction~\cite{qi2017pointnet++, mao2019interpolated} to obtain the point features for box prediction. 3DSSD~\cite{yang20203dssd} introduces Feature-FPS as a new sampling strategy for raw point clouds. Point-GNN~\cite{pointgnn} proposes a graph operator to aggregate the points information for object detection. The voxel-based single-stage detectors typically rasterize point clouds into voxel-grids and then apply 2D and 3D CNN to generate 3D proposals. VoxelNet~\cite{zhou2018voxelnet} divides points into voxels and leverages a 3D CNN to aggregate voxel features for proposal generation. SECOND~\cite{yan2018second} improves the voxel feature learning process by introducing 3D sparse convolutions. CenterPoints~\cite{yin2020center} proposes a center-based assignment that can be applied on feature maps for accurate location prediction. Pillar-based approaches generally transform point clouds into Bird-Eye-View (BEV) pillars and apply 2D CNNs for 3D object detection. PointPillars~\cite{lang2019pointpillars} is the first work that introduces the pillar representation. Pillar-based network~\cite{wang2020pillar} extends the idea by proposing the cylindrical view projection. Unlike the two-stage approaches, the single-stage methods cannot benefit from the fine-grained point information, which is crucial for accurate box prediction.

\noindent\textbf{Two-stage 3D object detection.} Two-stage approaches can be divided into $3$ streams, based on the representation of Points of Interest, \ie, point-based, voxel-based and point-voxel-based. Point-based approaches treat the sampled point clouds as Points of Interest. PointRCNN~\cite{shi2019pointrcnn} generates 3D proposals from raw point clouds and proposes Region Pooling to extract RoI features for the second stage refinement. STD~\cite{yang2019std} proposes a sparse-to-dense strategy and uses the PointsPool operation for RoI refinement. Voxel-based methods use the voxel points from 3D CNNs as Points of Interest. Part-$A^{2}$ Net~\cite{shi2020points} applies 3D sparse convolutions on the upsampled voxel points for RoI refinement. Voxel R-CNN~\cite{deng2020voxel} utilizes Voxel RoI Pooling to extract RoI features from voxels. Point-Voxel-based approaches use the keypoints that encode the whole scene as Points of Interest. PV-RCNN~\cite{shi2020pv} designs RoI-grid Pooling to aggregate keypoint features near RoIs. PV-RCNN++~\cite{shi2021pv} proposes Vector-Pooling to efficiently collect the keypoint features from different orientations. Compared with the previous methods, our Pyramid R-CNN shows better performance and robustness, and is compatible with all the representations of Points of Interest.

\begin{figure*}[!t]
\vspace{-6mm}
\centering
\includegraphics[width=1\textwidth]{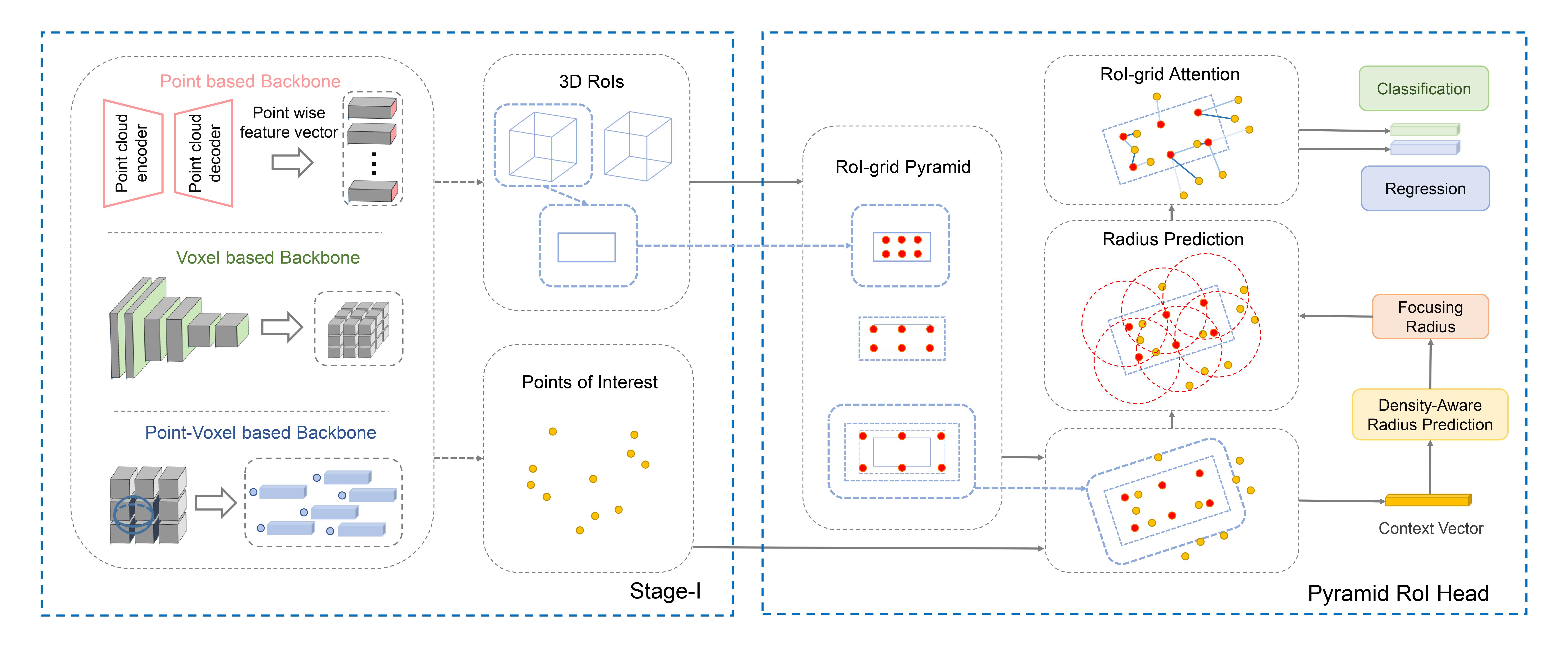}
\vspace{-5mm}
\caption{The overall architecture. Our Pyramid R-CNN can be plugged on diverse backbones (\eg point-based, voxel-based and point-voxel-based networks), which generate 3D proposals and Points of Interest (yellow points) on the stage-$1$. On the stage-$2$, we propose the pyramid RoI head that can be applied upon the 3D proposals and Points of Interest. In the pyramid RoI head, an RoI-grid Pyramid is first built to capture more context information. Then for each RoI-grid point (red point), a focusing radius $r$ (red dashed circle) is learned by the Density-Aware Radius Prediction module. Finally, RoI-grid Attention is performed on the Points of Interest within $r$ for box refinement.}
\label{fig_framework}
\vspace{-2mm}
\end{figure*}

\section{Pyramid R-CNN}
In this section, we detail the design of Pyramid R-CNN, a general two-stage framework for 3D object detection. We first introduce the overall architecture in \ref{Overall Architecture}. Then we introduce three key components in the pyramid RoI head: RoI-grid Pyramid in \ref{RoI-grid Pyramid}, RoI-grid Attention in \ref{RoI-grid Attention}, and the Density-Aware Radius Prediction (DARP) module in \ref{Density-Aware Radius Prediction}.

\subsection{Overall Architecture} \label{Overall Architecture}
Here, we present a new two-stage framework for accurate and robust 3D object detection, named Pyramid R-CNN, as shown in Figure~\ref{fig_framework}. The framework can be compatible with multiple backbones, \eg the point-based backbone, the voxel-based backbone , or the point-voxel-based backbone. On the first stage, those backbones output 3D proposals and corresponding Points of Interest: \eg point clouds near RoIs in~\cite{shi2019pointrcnn}, upsampled voxels in~\cite{shi2020points}, and keypoints in~\cite{shi2020pv}. On the second stage, we propose a novel pyramid RoI head, which consists of three key components: the RoI-grid Pyramid, RoI-grid Attention, and the Density-Aware Radius Prediction (DARP) module. For each RoI, we first build an RoI-grid Pyramid, by gradually enlarging the size of the original RoI in each pyramid level, and the coordinates of RoI-grid points are determined by the enlarged RoI size and the grid size. In each pyramid level, the focusing radius $r$ of the RoI-grid points is predicted from the global context vector through the Density-Aware Radius Prediction module. Then RoI-grid Attention parameterized by $r$ is performed to aggregate the features of Points of Interest into the RoI-grids. Finally, the RoI-grid features are enhanced and fed into two individual heads for classification and regression. We will describe those key components in the following sections.

\subsection{RoI-grid Pyramid} \label{RoI-grid Pyramid}

In this section, we present the RoI-grid Pyramid, a simple and effective module that captures rich context while still maintains internal structural information. Different from 2D feature pyramid~\cite{lin2017feature} which hierarchically encodes context information upon dense backbone features, our RoI-grid Pyramid is applied on each RoI by gradually placing the grid points \textit{out of} RoIs in a pyramid manner. The idea behind this design is based on the observation that image features inside RoIs generally contain sufficient semantic contexts, while point clouds inside RoIs contain quite limited information since object points are naturally sparse and incomplete. Even though each point has a large receptive field, the sparse compositional 3D shapes inside RoIs are hard to be recognized. In the following parts we will introduce detailed formulations.

RoI feature extraction generally relies on an RoI-grid for each RoI, and RoI-grid points collect the features of adjacent pixels or neighboring Points of Interest in the 2D or 3D cases respectively. Supposing we have an RoI with $W, L, H$ as width, length, and height and $(x_{c}, y_{c}, z_{c})$ as the bottom left corner, in standard RoI-grid representation, the $(i,j,k)$ RoI-grid point location $p^{ijk}_{grid}$ can be computed as:
\begin{equation}
    p^{ijk}_{grid} = (\frac{W}{N_{w}}, \frac{L}{N_{l}}, \frac{H}{N_{h}}) \cdot (0.5 + (i, j, k)) + (x_{c}, y_{c}, z_{c}),    
\end{equation}
where $(N_{w}, N_{l}, N_{h})$ are the grid sizes in three dimensions and all grid points are generated inside RoIs. 

\begin{figure}[!t] \centering   
\subfigure[standard RoI-grid] {
 \label{pyramid_1}     
\includegraphics[width=0.45\columnwidth]{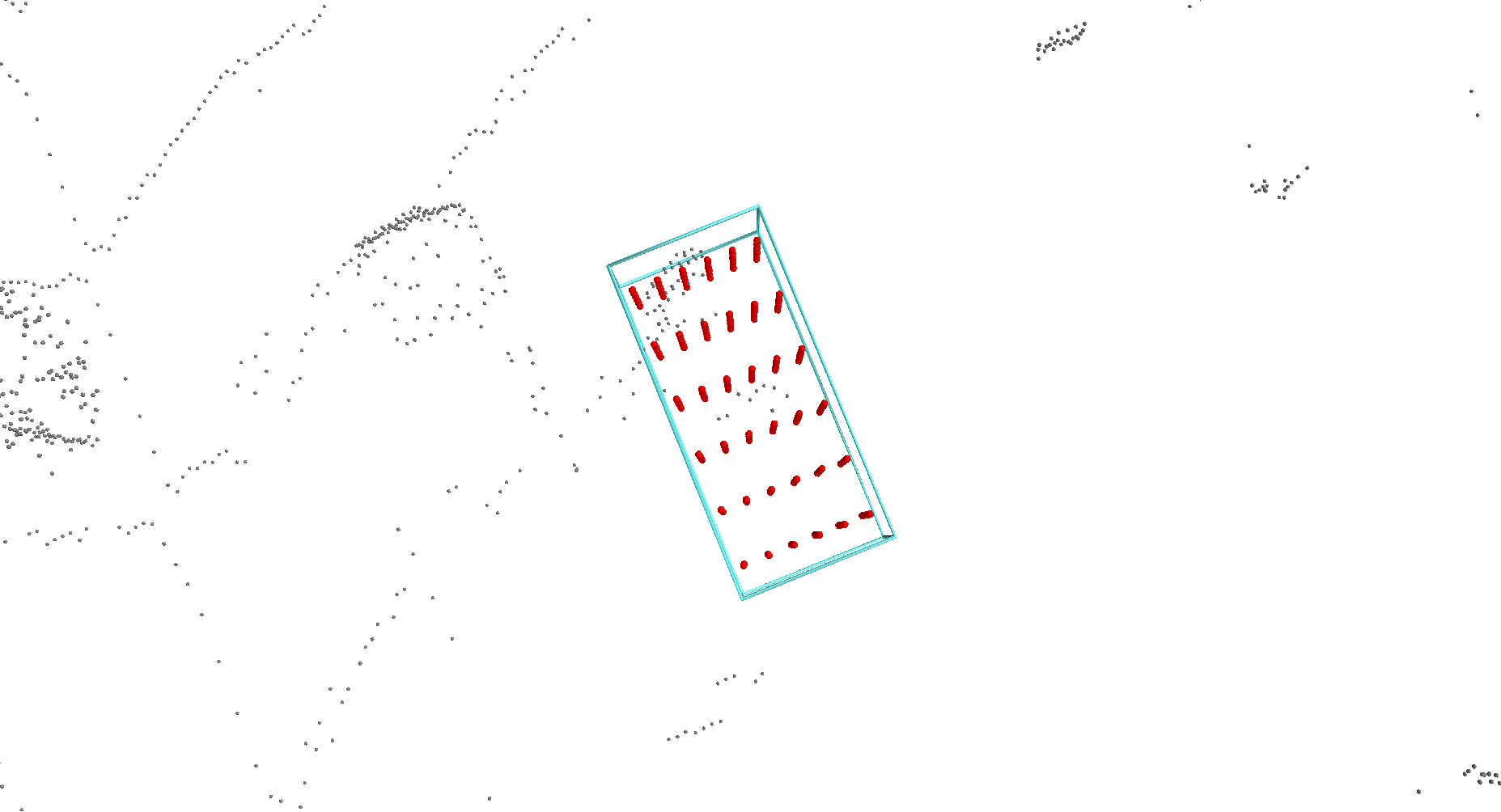}  
}    
\subfigure[RoI-grid Pyramid] { 
\label{pyramid_2}     
\includegraphics[width=0.45\columnwidth]{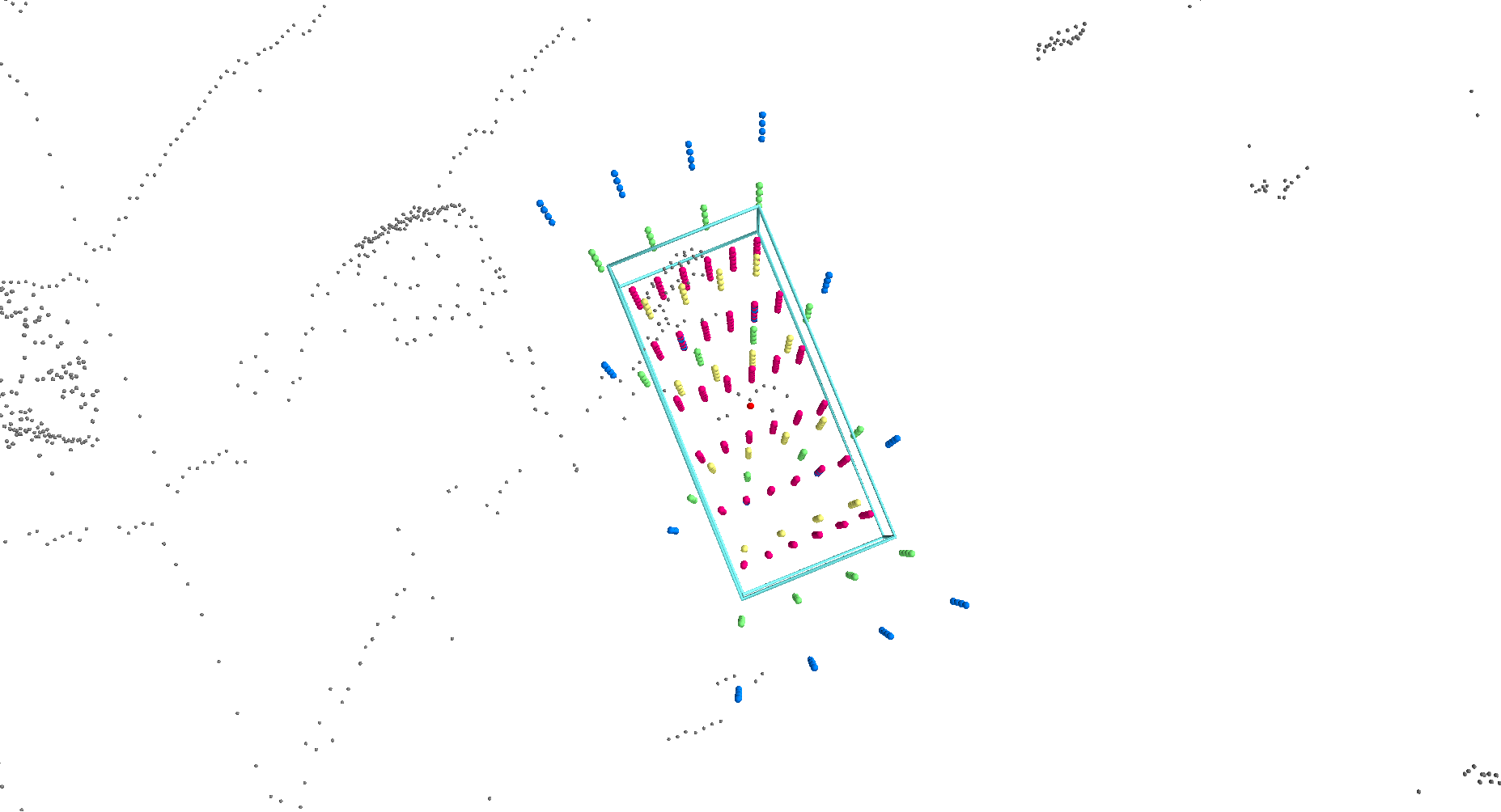}     
}
\subfigure[object/context points in (a)] { 
\label{pyramid_3}     
\includegraphics[width=0.45\columnwidth]{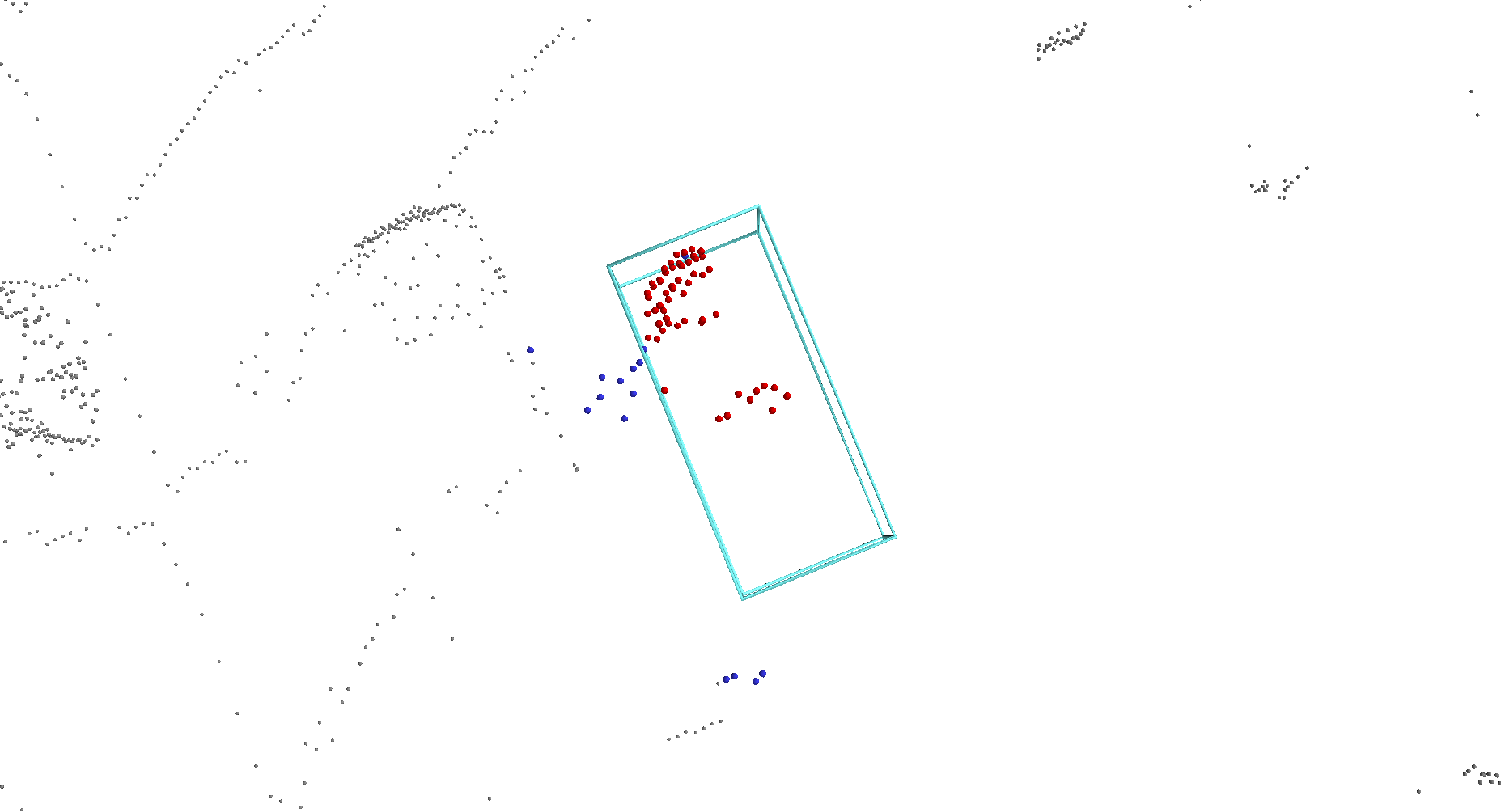}     
}
\subfigure[object/context points in (b)] { 
\label{pyramid_4}     
\includegraphics[width=0.45\columnwidth]{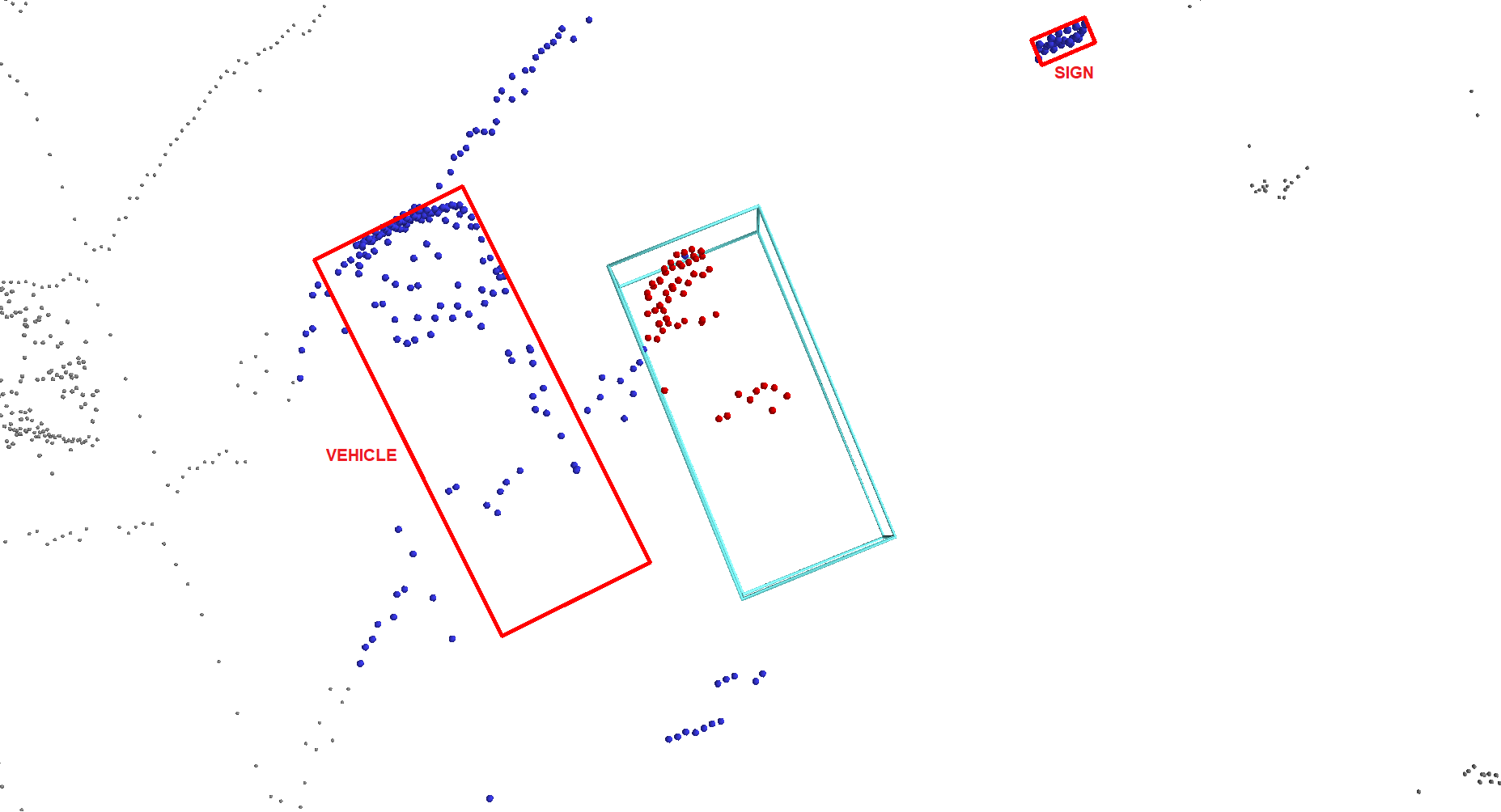}     
}  
\caption{Illustration of the RoI-grid Pyramid. Red points in \subref{pyramid_1} are the RoI-grid points, and different colors represent different pyramid levels in \subref{pyramid_2}. In \subref{pyramid_3} and \subref{pyramid_4} red points are object points and blue points are context points captured by the RoI. Compared to the standard RoI-grid, our RoI-grid Pyramid can capture more context points while maintain fine-grained internal structures, and by looking at neighboring vehicle and traffic sign (blue context points) outside the RoI, the cluster of red object points is easier to be recognized as a car.}     
\label{fig_pyramid} 
\vspace{-4mm}
\end{figure}

Utilizing features only inside the RoIs works well in the 2D detection models, mainly owing to two facts: the input feature map is dense and the collected pixels have large receptive fields. However, the cases are different in 3D models. As is shown in Figure~\ref{fig_pyramid}, the Points of Interest are naturally sparse and non-uniformly distributed inside the RoIs, and the object shape is extremely incomplete. Thus it is hard to accurately infer the sizes and categories of objects by solely collecting the features of few individual points and not referring to enough neighboring points information. 

To resolve the above problems, we propose the RoI-grid Pyramid which balances the fine-grained and large context information. The detailed structure is in Figure~\ref{fig_pyramid}. The key idea is to construct a pyramid grid structure that contains the RoI-grid points both inside and outside RoIs, so that the grid points inside RoIs can capture fine-grained shape structures for accurate box refinement, while the grid points outside RoIs can obtain large context information to identify incomplete objects. The grid points $p^{ijk}_{grid}$ for a pyramid level can be computed as:
\begin{equation}
    p^{ijk}_{grid} = (\frac{\rho_{w}W}{N_{w}^{\prime}}, \frac{\rho_{l}L}{N^{\prime}_{l}}, \frac{\rho_{h}H}{N^{\prime}_{h}}) \cdot (0.5 + (i, j, k)) + (x_{c}, y_{c}, z_{c}),
\end{equation}
where $\rho$ is the enlarging ratio of the original RoI size. $\rho$ starts from $1$ at the bottom level for maintaining fine-grained details, and becomes larger when the level goes higher to capture more context information. The grid size $N^{\prime}$ is initialized with the same value as the original $N$ at the bottom level and gets smaller at higher levels to save computational resources. For each pyramid level, features of grid points $f_{grid}$ are then aggregated by RoI-grid Attention from the features of Points of Interest. Finally, features of all pyramid levels are combined for boxes refinement. 

%Compared with conventional 2D counterparts which perform feature pyramids on backbones~\cite{lin2017feature}, we found that applying the pyramid structure on RoIs is more crucial considering the sparse characteristic of point clouds. And the RoI-grid Pyramid significantly enhances the detection accuracy of objects with only a few points with manageable computational resources.

\subsection{RoI-grid Attention} \label{RoI-grid Attention}
In this section, we introduce RoI-grid Attention, a novel RoI feature extraction operation that combines the state-of-the-art graph-based and attention-based point operators~\cite{wang2019dynamic, vaswani2017attention, zhao2020point} into a unified framework, and RoI-grid Attention can serve as a better substitute for conventional pooling-based operations~\cite{shi2020pv, deng2020voxel, shi2021pv} in 3D detection models. We first discuss the formulas of pooling-based, graph-based and attention-based point operators, and then we derive the formulation of RoI-grid Attention.

\textbf{Preliminary.} Let $p_{grid}$ be the coordinate of an RoI-grid point, and $p_{i}$, $f_{i}$ be the coordinate and the corresponding feature vector of the $i_{th}$ Points of Interest near $p_{grid}$. RoI feature extraction operation aims to obtain the respective feature vector $f_{grid}$ of the RoI-grid point $p_{grid}$, using the information of neighboring $p_{i}$ and $f_{i}$. 

\textbf{Pooling-based Operators.} The pooling-based operators are extensively applied for RoI feature extraction in most two-stage 3D detection models~\cite{shi2020pv, deng2020voxel, shi2021pv}. The neighboring feature $f_{i}$ and the relative location $p_{i} - p_{grid}$ first go through a MLP layer to obtain the transformed feature vector: $V^{i} = MLP([f_{i}, p_{i} - p_{grid}])$, where $[\cdot]$ is the concatenation function, and then a maxpooling operation is applied upon all the transformed features $V$ to obtain the RoI-grid feature $f^{pool}_{grid}$:
\begin{equation} \label{3.1.1}
    f^{pool}_{grid} = \mathop{maxpool}\limits_{i \in \Omega(r)}(V^{i}),
\end{equation} 
where $\Omega(r)$ means Points of Interest within the fixed radius $r$ of the RoI-grid point $p_{grid}$. The pooling-based operators only focus on the maximum channel response and this results in a loss of much semantic and geometric information.

\textbf{Graph-based Operators.} Graph-based operators can model the grid points and Points of Interest as a graph. The graph node $i$ represents the transformed feature of $f_{i}$: $V^{i} = MLP(f_{i})$, and the edge $Q^{i}_{pos}$ can be formulated as a linear projection of the location differences between two nodes: $Q^{i}_{pos} = Linear(p_{i} - p_{grid})$. For the graph node of a grid point $p_{grid}$, the feature $f^{graph}_{grid}$ is collected from adjacent nodes by a weighted combination operation. Following the same notations as Eq.\ref{3.1.1}, the general formula can be represented as
\begin{equation} \label{3.1.2}
    f^{graph}_{grid} = \sum_{i \in \Omega(r)}W(Q^{i}_{pos})\odot V^{i},
\end{equation}
where the function $W(\cdot)$ projects the graph edge embedding into the scalar or vector weight space, and $\odot$ denotes either the Hadamard product, dot product or scalar-vector product between learned weights and graph nodes.

\textbf{Attention-based Operators.} Attention-based operators can also be applied upon the grid points and Points of Interest. $Q^{i}_{pos}$ in Eq.\ref{3.1.2} can be viewed as the query embedding from the grid point $p_{grid}$ to the point $p_{i}$. $V^{i}$ is the value embedding obtained from the feature $f_{i}$ as Eq.\ref{3.1.2}. The key embedding $K^{i}$ can be formulated as $K^{i} = Linear(f_{i})$. Thus standard attention can be formulated as
\begin{equation} \label{3.1.3}
    f^{atten}_{grid} = \sum_{i \in \Omega(r)}W(Q^{i}_{pos}K^{i})\odot V^{i}.
\end{equation}
Additional normalization function, \ie softmax, is applied in $W(\cdot)$. Recently proposed Point Transformer~\cite{zhao2020point} extending the idea of standard attention and the formula can be represented as
\begin{equation} \label{3.1.4}
    f^{tr}_{grid} = \sum_{i \in \Omega(r)}W(K^{i} + Q^{i}_{pos})\odot (V^{i} + Q^{i}_{pos}).
\end{equation}

\begin{figure}[!t]
\centering
\includegraphics[width=0.5\textwidth]{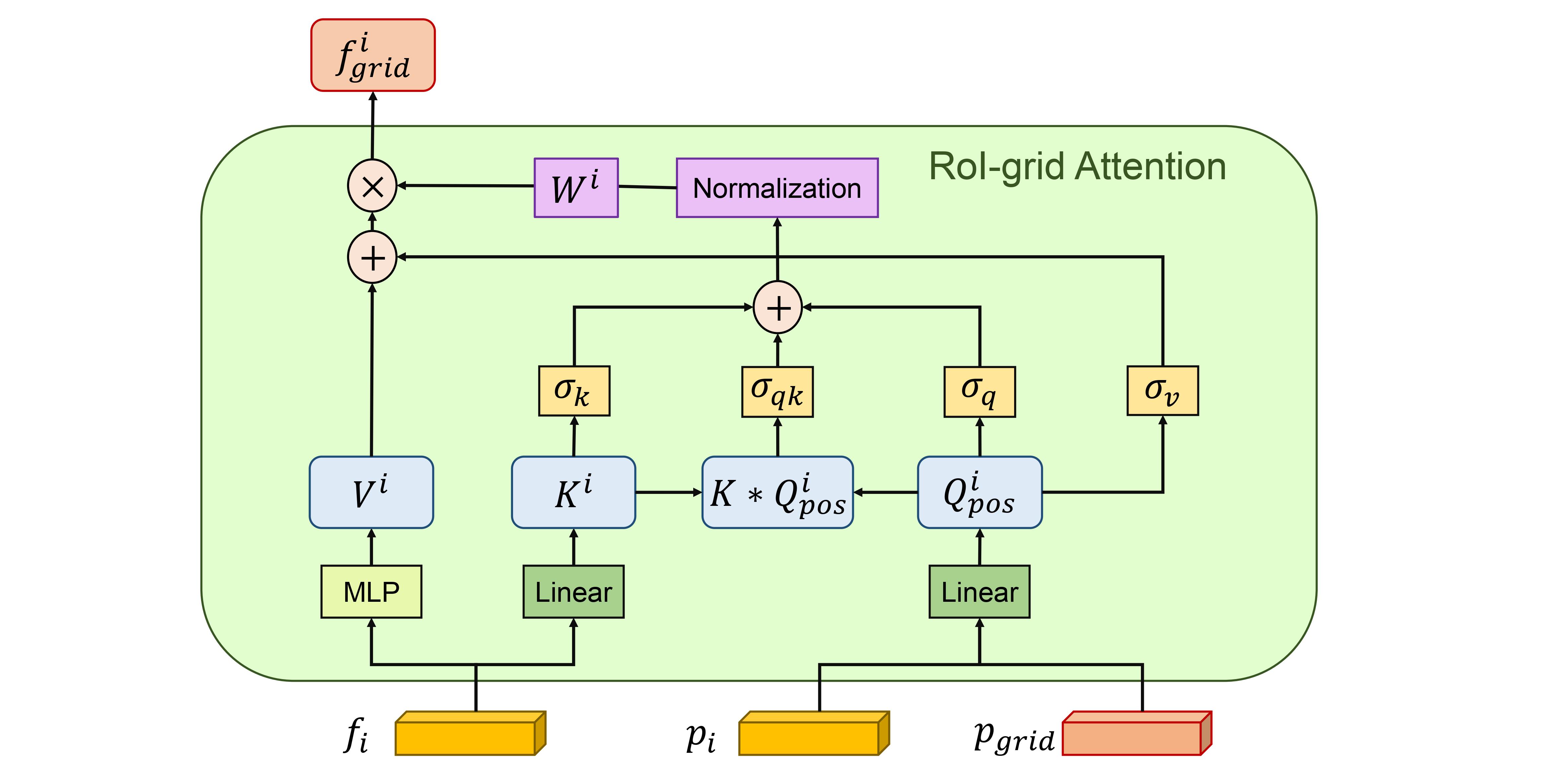}
\caption{Illustration of RoI-grid Attention. RoI-grid Attention introduces learnable gated functions $\sigma_{\star}$ to dynamically select the attention components, and it provides a unified formulation that includes the conventional graph and attention operators.}
\label{fig_attention}
\vspace{-4mm}
\end{figure}

\textbf{RoI-grid Attention.} In our approach, we analyze the structural similarity of Eq.\ref{3.1.2}, Eq.\ref{3.1.3} and Eq.\ref{3.1.4}. We find that those formulas have common basic elements and operators. Thus it's natural to merge those formulas into a unified framework with gated functions. We name this new formula RoI-grid Attention:
\begin{equation} \label{3.1.5}
\begin{aligned}
f_{grid} = \sum_{i \in \Omega(r)} & W(\sigma_{k} K^{i} + \sigma_{q} Q^{i}_{pos} + \sigma_{qk} Q^{i}_{pos}K^{i})\\& \odot (V^{i} + \sigma_{v} Q^{i}_{pos}),
\end{aligned}
\end{equation}
where $\sigma_{\ast}$ is a learnable gated function which can be implemented by a linear projection of the respective embedding with a sigmoid activation output. RoI-grid Attention is a generalized formulation combining graph-based and attention-based operations. We can derive the graph operator Eq.\ref{3.1.2} from Eq.\ref{3.1.5} when $\sigma_{q}$, $\sigma_{k}$, $\sigma_{qk}$, $\sigma_{v}$ are $1$, $0$, $0$, $0$ respectively. Similarly, we can derive the standard attention Eq.\ref{3.1.3} when $\sigma_{q}$, $\sigma_{k}$, $\sigma_{qk}$, $\sigma_{v}$ are $0$, $0$, $1$, $0$, or Point Transformer Eq.\ref{3.1.4} when $\sigma_{q}$, $\sigma_{k}$, $\sigma_{qk}$, $\sigma_{v}$ are $1$, $1$, $0$, $1$.

RoI-grid Attention is a flexible and effective operation for RoI feature extraction. With the learnable gated functions, RoI-grid Attention is able to learn which point is significant to the RoI-grid points, from both the geometric information $Q_{pos}$ and the semantic information $K$, as well as their combinations $Q_{pos}K$ adaptively. With $\sigma_{v}$, RoI-grid Attention can also learn to balance the ratio of geometric features $Q_{pos}$ and semantic features $V$ used in feature aggregation. Compared with the pooling-based methods, only a few linear projection layers are added in RoI-grid Attention, which maintains the computational efficiency. Replacing pooling-based operators with RoI-grid Attention consistently boosts the detection performance.

\subsection{Density-Aware Radius Prediction} \label{Density-Aware Radius Prediction}
In this section, we investigate the learning problem of the radius $r$, which determines the range $\Omega(r)$  of neighboring Points of Interest that participate in the feature extraction process. The radius $r$ is a hyper-parameter used in all the point operators in~\ref{RoI-grid Attention}, and has to be determined by researchers in previous approaches. The fixed and predefined $r$ cannot adapt to the density changes of point clouds, and may lead to empty spherical ranges if not set properly. In this paper, we make the prediction of $r$ a fully-differentiable process and further propose the Density-Aware Radius Prediction (DARP) module, aiming at learning an adaptive neighborhood for RoI feature extraction. We first introduce the general formulation of RoI-grid Attention from a probabilistic perspective. Next, we propose a novel method to differentiate the learning of $r$. Finally, we introduce the design of the DARP module.

RoI-grid Attention is composed of two steps: first selects Points of Interest within the radius $r$, and next performs weighted combinations on those points. With the same notations in \ref{RoI-grid Attention}, we can reformulate the first step as sampling from a conditional distribution $p(i|r)$:
\begin{equation} \label{3.2.1}
    p(i|r) = 
    \begin{cases}
    0 & ||p_{i} - p_{grid}||_{2} > r\\
    1 & ||p_{i} - p_{grid}||_{2} \leq r
    \end{cases}
\end{equation}
Then the second step can be represented as calculating the probabilistic expectation:
\begin{equation} \label{3.2.2}
    f_{grid} = \mathbb{E}_{i \sim p(i|r)}[W^{i} \odot V^{i}],
\end{equation}
where $W^{i}$ denotes $W(\sigma_{k} K^{i} + \sigma_{q} Q^{i}_{pos} + \sigma_{qk} Q^{i}_{pos}K^{i})$ and $V^{i}$ denotes $(V^{i} + \sigma_{v} Q^{i}_{pos})$ with a slight abuse of notations. 

We propose a new probability distribution $s(i|r)$ as a substitute for $p(i|r)$, and $s(i|r)$ should satisfy two requirements: \romannumeral1) $s(i|r)$ should have similar characteristics as $p(i|r)$, which means that most points sampled from $s(i|r)$ should be inside $r$; \romannumeral2) $s(i|r)$ should also leave a few points outside $r$, mainly for the exploration of the surrounding environment. Thus we formulate the probability $s(i|r)$ as:
\begin{equation} \label{3.2.3}
    s(i|r) = 1 - sigmoid(\frac{||p_{i} - p_{grid}||_{2} - r}{\tau}),
\end{equation}
where $sigmoid(x) = (1 + e^{-x})^{-1}$ and $\tau$ is the temperature which controls the decay rate of probability. With a small $\tau$, $s(i|r)$ is close to $1$ when $p_{i}$ is inside $r$, and is close to $0$ if outside, while near the spherical boundary the sampling probability $s(i|r)$ is between $0$ and $1$. With $s(i|r)$ as a smooth approximation to $p(i|r)$, we want to compute the gradient of $r$ from the approximated RoI-grid Attention:
\begin{equation} \label{3.2.4}
    \nabla_{r} f_{grid} = \nabla_{r} \mathbb{E}_{i \sim s(i|r)}[W^{i} \odot V^{i}].
\end{equation}

However, taking the derivative \wrt $r$ is still infeasible, since we cannot directly calculate the gradient of a parameterized distribution. The reparameterization trick~\cite{kingma2013auto} offers a possible solution to the problem. The key insight is sampling from a basic distribution and then move the original distribution parameters inside the expectation function as coefficients. The gradient of $r$ can be computed as:
\begin{equation} \label{3.2.5}
    \nabla_{r} f_{grid} = \mathbb{E}_{i \sim U(\epsilon)}[\nabla_{r} [s(i, r) \cdot W^{i} \odot V^{i}]],
\end{equation}
where $s(i, r)$ is the same as Eq.\ref{3.2.3}, and the theoretical distribution $U(\epsilon)=1$ means that the sampling probability is $1$ in the whole 3D space. In practical, considering the fact that $s(i, r)$ is close to $0$ when $\epsilon \gg r$, we apply an approximation and restrict the sampling range $U(\epsilon)$ within a sphere with a radius slightly larger than $r$, \ie $r+5\tau$ in our experiments. This approximation reduces the computational overhead to the same level as vanilla RoI-grid Attention. Since $s(i, r)$ is a differentiable function \wrt $r$, we are able to compute the gradient of $r$ in a differential manner using Eq.\ref{3.2.5}. The new formulation of RoI-grid Attention can be represented as \begin{equation} \label{3.2.6}
\begin{aligned}
f_{grid} = \sum_{i \in U(\epsilon)} & W(\sigma_{k} K^{i} + \sigma_{q} Q^{i}_{pos} + \sigma_{qk} Q^{i}_{pos}K^{i})\\& \odot (V^{i} + \sigma_{v} Q^{i}_{pos}) \cdot s(i, r).
\end{aligned}
\end{equation}
Compared with vanilla RoI-grid Attention in Eq.\ref{3.1.5}, a slightly larger sampling range $r + 5\tau$ is used and an coefficient $s(i, r)$ is added into the original formula, which costs little additional resources. Although several approximations are applied, we found that they didn't hamper the training but boost the performance in our experiments. 

\begin{figure}[!t]
\centering
\includegraphics[width=0.45\textwidth]{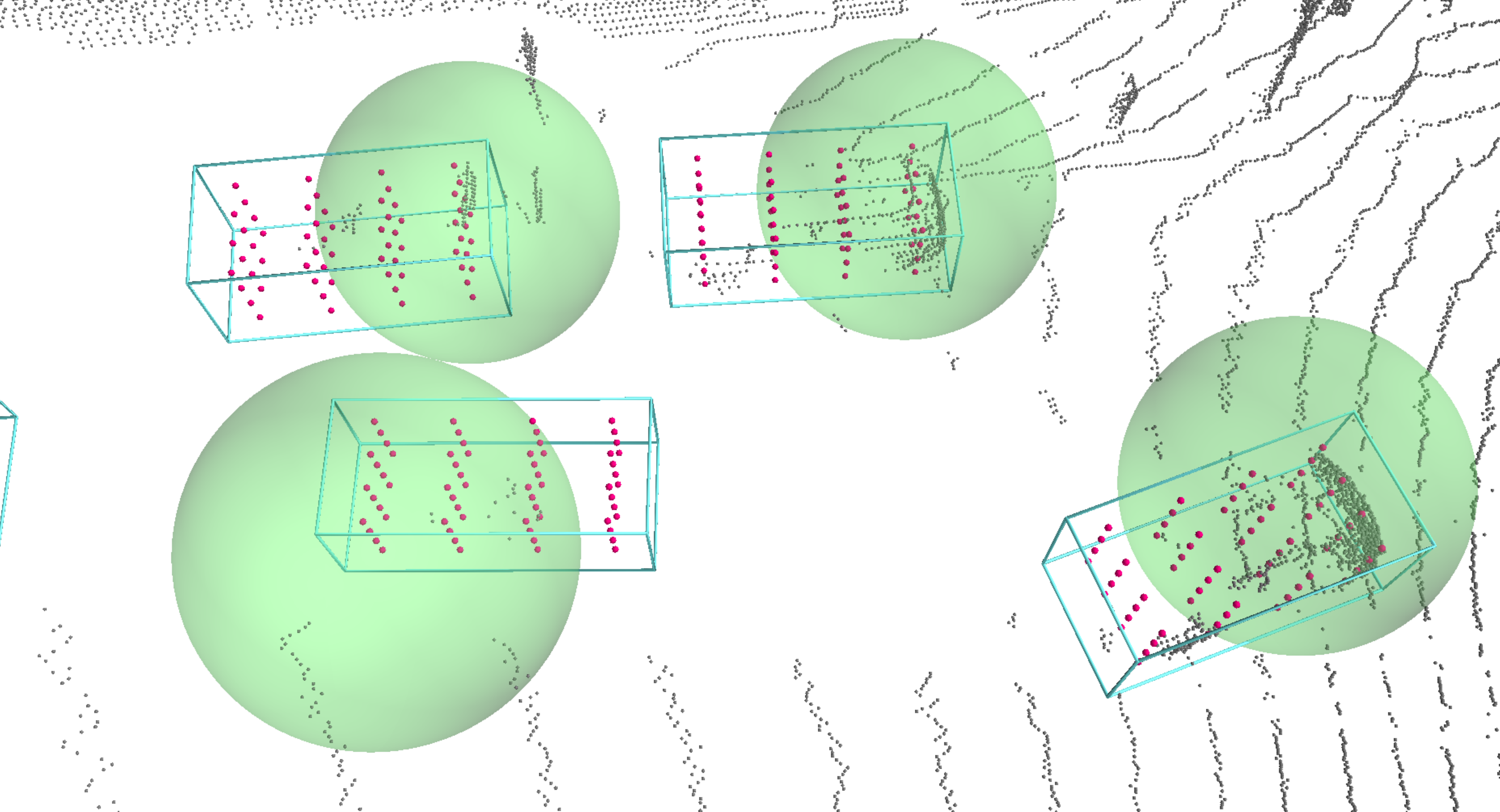}
\caption{Illustration of dynamic radius predicted by the Density-Aware Radius Prediction module. For each RoI, an adaptive focusing radius is learned based on the sparsity conditions.}
\label{fig_radius}
\vspace{-4mm}
\end{figure}

We further propose the DARP module based on Eq.\ref{3.2.6}. For each pyramid level, a context embedding is obtained by summarizing the information of Points of Interest near this RoI, and then the embedding is utilized to predict the radius $r$ for all grid points in this level. $r$ is further transformed into an coefficient by $s(i, r)$ and participates in the computation of RoI-grid Attention. Since the context embedding captures point cloud information, \ie density, shape, \etc, the predicted $r$ is able to adapt to the environmental changes, and is more robust than the human-defined counterpart.

\section{Experiments}
In this section, we evaluate our Pyramid R-CNN on the commonly used Waymo Open dataset~\cite{sun2020scalability} and the KITTI~\cite{geiger2013vision} dataset. We first introduce the experimental settings in \ref{Experimental Setup} and then compare our approach with previous state-of-the-art methods on the Waymo Open dataset in \ref{3D Detection on the Waymo Open Dataset} and the KITTI dataset in \ref{3D Detection on the KITTI Dataset}. Finally, we conduct ablation studies to evaluate the efficacy of each component in \ref{Ablation Studies}.

\subsection{Experimental Setup} \label{Experimental Setup}

\begin{table*}[!t]
\setlength{\tabcolsep}{4.1mm}{
\begin{tabular}{l|c|c|c c c}
\toprule
\multirow{2}{*}{Methods} & \multirow{1}{*}{LEVEL\_1} & \multirow{1}{*}{LEVEL\_2} & \multicolumn{3}{c}{LEVEL\_1 3D mAP/mAPH by Distance} \\
 &   3D mAP/mAPH           &       3D mAP/mAPH      
 & 0-30m            & 30-50m          & 50m-Inf         \\
\midrule
PointPillars~\cite{lang2019pointpillars}   & 63.3/62.7  & 55.2/54.7  & 84.9/84.4  & 59.2/58.6    & 35.8/35.2       \\
MVF~\cite{MVF}  & 62.93/-      & -      & 86.30/-     & 60.02/-         & 36.02/-         \\
Pillar-OD~\cite{wang2020pillar}     & 69.8/-   & -   & 88.5/-    & 66.5/-    & 42.9/-     \\
AFDet~\cite{ge2020afdet}  & 63.69/-    &  -  & 87.38/-  & 62.19/-    & 29.27/-         \\
LaserNet~\cite{meyer2019lasernet}  & 52.1/50.1  & -   & 70.9/68.7   & 52.9/51.4  & 29.6/28.6 \\
CVCNet~\cite{chen2020every}  & 65.2/-  & -   & 86.80/-    & 62.19/-    & 29.27/-         \\
StarNet~\cite{ngiam2019starnet} & 64.7/56.3   & 45.5/39.6 & 83.3/82.4& 58.8/53.2 & 34.3/25.7 \\
RCD~\cite{bewley2020range} & 69.0/68.5  & - & 87.2/86.8  & 66.5/66.1    & 44.5/44.0       \\
Voxel R-CNN~\cite{deng2020voxel}  & 75.59/-  & 66.59/-  & 92.49/-    & 74.09/-   & 53.15/-  \\
\midrule
PointRCNN$^{\star}$~\cite{shi2019pointrcnn}   & 45.05/44.25 & 37.41/36.74  & 72.24/71.31 &       31.21/30.41 & 23.77/23.15   \\
\textbf{Pyramid-P (ours)}    & \textbf{47.02/46.58} & \textbf{39.10/38.76} & \textbf{74.24/73.78}  & \textbf{32.49/31.96}  &   \textbf{25.68/25.24}     \\
\midrule
Part-$A^{2}$ Net$^{\star}$~\cite{shi2020points}   &  71.69/71.16 & 64.21/63.70  & 91.83/91.37     & 69.99/69.37  & 46.26/45.41  \\
\textbf{Pyramid-V (ours)}    & \textbf{75.83/75.29} & \textbf{66.77/66.28} 
     & \textbf{92.63/92.20}   & \textbf{74.46/73.84}    & \textbf{53.40/52.44}           \\
\midrule
PV-RCNN~\cite{shi2020pv}                  & 70.3/69.7                           & 65.4/64.8                           & 91.9/91.3        & 69.2/68.5       & 42.2/41.3       \\
\textbf{Pyramid-PV (ours)}   & \textbf{76.30/75.68}  & \textbf{67.23/66.68}                         & \textbf{92.67/92.20}      & \textbf{74.91/74.21}   & \textbf{54.54/53.45}     \\
%\textbf{Pyramid-PV$^{\dag}$ (ours)}  & \textbf{78.70/78.12} & \textbf{69.70/69.17}                  & \textbf{92.83/92.37}      & \textbf{75.61/75.01}    & \textbf{57.44/56.53}     \\
%\textbf{Pyramid-PV$^{\ddag}$ (ours)}     & \textbf{79.72/79.12}  & \textbf{70.85/70.36}             & \textbf{92.84/92.38}      & \textbf{78.60/78.03}     & \textbf{59.81/59.02}     \\
\bottomrule
\end{tabular}}
\setlength{\belowcaptionskip}{10pt}
\caption{Performance comparison on the Waymo Open Dataset with 202 validation sequences for the vehicle detection. $\star$: re-implemented by ourselves with the official code.} \label{table_waymo_1}
\vspace{-2mm}
\end{table*}

\begin{table*}[]
\setlength{\tabcolsep}{3.95mm}{
\begin{tabular}{l|c|c|c c c}
\toprule
\multirow{2}{*}{Methods} & \multirow{1}{*}{LEVEL\_1} & \multirow{1}{*}{LEVEL\_2} & \multicolumn{3}{c}{LEVEL\_1 3D mAP/mAPH by Distance} \\
 &   3D mAP/mAPH           &       3D mAP/mAPH      
 & 0-30m            & 30-50m          & 50m-Inf         \\
\midrule
CenterPoint$^{\star}$~\cite{yin2020center}  & 81.05/80.59 & 73.42/72.99  & 92.52/92.13 & 79.94/79.43     & 61.06/60,42     \\
PV-RCNN$^{\star}$~\cite{shi2020pv} & 81.06/80.57 & 73.69/73.23 & 93.40/92.98  & 80.12/79.57   
& 61.22/60.47     \\
\textbf{Pyramid-PV$^{\ddag}$ (ours)} & \textbf{81.77/81.32} & \textbf{74.87/74.43}                         & \textbf{93.19/92.80}   & \textbf{80.53/80.04}   & \textbf{64.55/63.84} \\  
\bottomrule
\end{tabular}}
\setlength{\belowcaptionskip}{10pt}
\caption{Performance comparison on the Waymo Open Dataset \textit{test} leaderboard for the vehicle detection. $\star$: test submissions are the modified version of original architectures. $\ddag$: We append another frame following~\cite{shi2020pv} and use a larger voxel backbone.} \label{table_waymo_2}
\vspace{-4mm}
\end{table*}

\noindent\textbf{Waymo Open Dataset.} The Waymo Open Dataset contains $1000$ sequences in total, including $798$ sequences (around $158k$ point cloud samples) in the training set and 202 sequences (around $40k$ point cloud samples) in the validation set. The official evaluation metrics are standard 3D mean Average Precision (mAP) and mAP weighted by heading accuracy (mAPH). Both of the two metrics are based on an IoU threshold of 0.7 for vehicles and 0.5 for other categories. The testing samples are split in two ways. The first way is based on the distances of objects to the sensor: $0-30m$, $30-50m$ and $>50m$. The second way is according to the difficulty levels: LEVEL\_1 for boxes with more than five LiDAR points and LEVEL\_2 for boxes with at least one LiDAR point.

\noindent\textbf{KITTI Dataset.} The KITTI dataset contains $7481$ training samples and $7518$ test samples, and the training samples are further divided into the \textit{train} split ($3712$ samples) and the $val$ split ($3769$ samples). The official evaluation metric is mean Average Precision (mAP) with a rotated IoU threshold 0.7 for cars. On the \textit{test} set mAP is calculated with $40$ recall positions by the official server. The results on the \textit{val} set are calculated with 11 recall positions for a fair comparison with other approaches.

We provide $3$ architectures of Pyramid R-CNN, compatible with the point-based, the voxel-based and the point-voxel-based backbone, respectively. We would like readers to refer to~\cite{openpcdet} for the detailed design of those backbones. 

\noindent\textbf{Pyramid-P.} \textit{\textbf{Pyramid} R-CNN for \textbf{P}oints} is built upon the point-based method PointRCNN~\cite{shi2019pointrcnn}. In particular, we replace the Canonical 3D Box Refinement module of PointRCNN, with our proposed pyramid RoI head in Pyramid R-CNN, and we still use the sampled points in~\cite{shi2019pointrcnn} as Points of Interest. The point cloud backbone and other configurations are kept the same for a fair comparison.

\noindent\textbf{Pyramid-V.} \textit{\textbf{Pyramid} R-CNN for \textbf{V}oxels} is built upon the voxel-based method Part-$A^{2}$ Net~\cite{shi2020points}. Specifically, we replace the 3D sparse convolutional head of Part-$A^{2}$ Net, with our proposed pyramid RoI head in Pyramid R-CNN, and we still use the upsampled voxels as Points of Interest. The voxel-based backbone and other configurations are kept the same for a fair comparison.

\noindent\textbf{Pyramid-PV.} \textit{\textbf{Pyramid} R-CNN for \textbf{P}oint-\textbf{V}oxels} is designed upon the point-voxel-based method PV-RCNN~\cite{shi2020pv}. In particular, we replace the RoI-grid Pooling module of PV-RCNN, with our proposed pyramid RoI head in Pyramid R-CNN, and we still use the keypoints as Points of Interest. The keypoints encoding process, the 3D sparse convolutional networks and other configurations are kept the same for a fair comparison.

\noindent\textbf{Implementation Details.} Here we only introduce the architecture of Pyramid-PV on the Waymo Open dataset. The implementations of other models are similar and can be found in the supplementary materials. In RoI-grid Attention, the number of attention heads is set to $4$ and each head contains $16$ feature channels. In the DARP module, the context embedding is extracted from the neighboring Points of Interest within two spheres with the radius $2.4m$ and $4.8m$. The temperature $\tau$ starts from $0.02$ and exponentially decays to $0.0001$ in the end. The RoI-grid Pyramid consists of $5$ levels, with the number of grid points as $6^{3}, 4^{3}, 4^{3}, 4^{3}, 1$ respectively, and for each pyramid level, a focusing radius $r$ is predicted and shared across all the grid points in this level. The enlarging ratio $\rho_{w}$ and $\rho_{l}$ are set to $1, 1, 1.5, 2, 4$ for the respective level of the RoI-grid Pyramid, and $\rho_{h}$ is set to $1$ in all pyramid levels. The maximum number of points that participate in RoI-grid Attention for each grid point is set to $8, 16, 16, 16, 32$ for the corresponding pyramid level.

%\vspace{-1mm}
\noindent\textbf{Training and Inference Details.} Our Pyramid R-CNN is trained from scratch with the ADAM optimizer. On the KITTI dataset, Pyramid-P, Pyramid-V and Pyramid-PV are trained with the same batch size $16$, the learning rate $0.01, 0.01, 0.005$ respectively for $80$ epochs on $8$ V100 GPUs. On the Waymo Open dataset, we uniformly sample $20\%$ frames for training and use the full validation set for evaluation following~\cite{shi2020pv}. Pyramid-P, Pyramid-V and Pyramid-PV are trained with the same batch size $32$, the learning rate $0.01$ for $40$ epochs. The cosine annealing learning rate strategy is adopted for the learning rate decay. Other configurations are kept the same as the corresponding baselines~\cite{shi2019pointrcnn, shi2020points, shi2020pv} for a fair comparison.

\vspace{-1mm}
\subsection{Comparisons on the Waymo Open Dataset} \label{3D Detection on the Waymo Open Dataset}
\vspace{-1mm}

We evaluate the performance of Pyramid R-CNN on the Waymo Open dataset. The validation results in Table~\ref{table_waymo_1} show that our Pyramid-P, Pyramid-V and Pyramid-PV significantly outperform the baseline methods with $2.0\%$, $4.1\%$ and $6.0\%$ mAP gain respectively, and achieves superior mAP on all difficulty levels and all distance ranges, which demonstrates the effectiveness and generalizability of our approach. It is worth noting that Pyramid-V surpasses PV-RCNN by $12.3\%$ mAP in detecting objects that are $>50m$, which indicates the adaptability of our approach to the extremely sparse conditions. Our Pyramid-PV outperforms all the previous approaches with a remarkable margin, and achieves the new state-of-the-art performance $76.30\%$ mAP and $67.23\%$ mAP for the LEVEL\_1 and LEVEL\_2 difficulty. In table~\ref{table_waymo_2}, our Pyramid-PV$^{\ddag}$ achieves $81.77\%$ LEVEL\_1 mAP, ranks $1^{st}$ on the Waymo vehicle detection leaderboard as of March 10th, 2021, and surpasses all the LiDAR-only approaches.

\subsection{Comparisons on the KITTI Dataset} \label{3D Detection on the KITTI Dataset}
We evaluate our Pyramid R-CNN on the KITTI dataset. The \textit{test} results in Table~\ref{table_kitti_1} show that our Pyramid-P, Pyramid-V and Pyramid-PV consistently outperform the baseline methods with $4.66\%$, $2.79\%$ and $0.65\%$ mAP gain respectively on the moderate car class, and Pyramid-PV achieves $82.08\%$ mAP, becoming the new state-of-the-art. The validation results in Table~\ref{table_kitti_2} show that Pyramid-P, Pyramid-V and Pyramid-PV improve the baselines by $4.47\%$, $3.67\%$ and $0.69\%$ mAP on the moderate car class, and $1.06\%$, $0.07\%$ and $0.14\%$ mAP on the hard car class respectively. We note that the performance gains are mainly from the hard cases, which indicates the adaptability of our approach, and the observations on the KITTI dataset are consistent with those on the Waymo Open dataset. 

\begin{table}[]
\setlength{\tabcolsep}{1.8mm}{
\begin{tabular}{|l|c| c c c|}
\toprule
\multirow{2}{*}{Methods} & \multirow{2}{*}{Modality} & \multicolumn{3}{c}{$AP_{3D}$ (\%)} \\
                        &                           & Easy   & Mod.   & Hard   \\
\midrule
MV3D~\cite{mv3d}        & R+L                       & 74.97  & 63.63  & 54.00  \\
AVOD-FPN~\cite{avod}    & R+L                       & 83.07  & 71.76  & 65.73  \\
F-PointNet~\cite{fp}    & R+L                       & 82.19  & 69.79  & 60.59  \\
MMF~\cite{mmf}          & R+L                       & 88.40  & 77.43  & 70.22  \\
3D-CVF~\cite{3dcvf}     & R+L                       & 89.20  & 80.05  & 73.11  \\
CLOCs~\cite{clocs}      & R+L                       & 88.94  & 80.67  & 77.15  \\
ContFuse~\cite{cont}    & R+L                       & 83.68  & 68.78  & 61.67  \\
\midrule
VoxelNet~\cite{zhou2018voxelnet}    & L             & 77.47  & 65.11  & 57.73  \\
PointPillars~\cite{lang2019pointpillars}    & L     & 82.58  & 74.31  & 68.99  \\
SECOND~\cite{yan2018second}     & L                 & 84.65  & 75.96  & 68.71  \\
STD~\cite{yang2019std}          & L                 & 87.95  & 79.71  & 75.09  \\
Patches~\cite{patch}    & L                         & 88.67  & 77.20  & 71.82  \\
3DSSD~\cite{yang20203dssd}      & L                 & 88.36  & 79.57  & 74.55  \\
SA-SSD~\cite{he2020structure}   & L                 & 88.75  & 79.79  & 74.16  \\
TANet~\cite{tanet}      & L                         & 85.94  & 75.76  & 68.32  \\
Voxel R-CNN~\cite{deng2020voxel}    & L             & 90.90  & 81.62  & 77.06  \\
HVNet~\cite{ye2020hvnet}    & L                     & 87.21  & 77.58  & 71.79  \\
PointGNN~\cite{pointgnn}    & L                     & 88.33  & 79.47  & 72.29  \\
\midrule
PointRCNN~\cite{shi2019pointrcnn}   & L             & 86.96  & 75.64  & 70.70  \\
\textbf{Pyramid-P (ours)}    & L         & \textbf{87.03} & \textbf{80.30} & \textbf{76.48}  \\
\midrule
Part-$A^{2}$ Net~\cite{shi2020points} & L           & 87.81  & 78.49  & 73.51  \\
\textbf{Pyramid-V (ours)}       & L & \textbf{87.06}  & \textbf{81.28} & \textbf{76.85}\\
\midrule
PV-RCNN~\cite{shi2020pv}        & L                 & 90.25  & 81.43  & 76.82  \\
\textbf{Pyramid-PV (ours)}      & L & \textbf{88.39}  & \textbf{82.08}  & \textbf{77.49} \\ 
\bottomrule
\end{tabular}}
\setlength{\belowcaptionskip}{10pt}
\caption{Performance comparison on the KITTI \textit{test} set with AP calculated by $40$ recall positions for the car category. R+L denotes the methods that combines RGB data and point clouds. L denotes LiDAR-only approaches.} \label{table_kitti_1}
\vspace{-3mm}
\end{table}

\begin{table}[]
\setlength{\tabcolsep}{3.77mm}{
\begin{tabular}{|l| c c c|}
\toprule
\multirow{2}{*}{Methods} & \multicolumn{3}{c}{$AP_{3D}$ (\%)} \\
                                        & Easy   & Mod.   & Hard   \\
\midrule
PointRCNN~\cite{shi2019pointrcnn}       & 88.88  & 78.63  & 77.38  \\
\textbf{Pyramid-P (ours)}               & \textbf{88.47} & \textbf{83.10}  & \textbf{78.44} \\
\midrule
Part-$A^{2}$ Net~\cite{shi2020points}   & 89.47  & 79.47  & 78.54  \\
\textbf{Pyramid-V (ours)}               & \textbf{88.44} & \textbf{83.14}  & \textbf{78.61} \\
\midrule
PV-RCNN~\cite{shi2020pv}                & 89.35  & 83.69  & 78.70  \\
\textbf{Pyramid-PV (ours)}              & \textbf{89.37} & \textbf{84.38}  & \textbf{78.84} \\
\bottomrule
\end{tabular}}
\setlength{\belowcaptionskip}{10pt}
\caption{Performance comparison on the KITTI \textit{val} split with AP calculated by $11$ recall positions for the car category.} \label{table_kitti_2}
\vspace{-3mm}
\end{table}

\subsection{Ablation Studies} \label{Ablation Studies}
\textbf{The effects of different components.} As is shown in Table~\ref{table_ablation_1}, on the Waymo validation set, the RoI-grid Pyramid of the Pyramid-PV model improves over the baseline by $1.20\%$ mAP, mainly because the RoI-grid Pyramid is able to capture large context information which benefits the detection of the hard cases. Based on the RoI-grid Pyramid, replacing RoI-grid Pooling with RoI-grid Attention boosts the performance by $0.51\%$ mAP, which indicates that RoI-grid Attention is a more effective operation than RoI-grid Pooling. Using the adaptive radius $r$ instead of the fixed radius boosts the performance by $0.37\%$ mAP, which demonstrates the efficacy of the DARP module. 

\textbf{The effects of different pyramid configurations.} As is shown in Table~\ref{table_ablation_2}, we found that the RoI-grid Pyramid with $\rho_{w}, \rho_{l}>1$ enhances the performance compared with the standard RoI-grid only with $\rho_{w}, \rho_{l}=1$, mainly because placing some grid points outside RoIs encodes richer contexts. The total number of used grid points is $409$, which is comparable to $432$ grid points used in~\cite{shi2020pv}.

\textbf{Inference speed analysis.} We test the inference speed of different frameworks under a single V100 GPU with batch size $1$, and obtain the average running speed of all samples in KITTI \textit{val} split. Table~\ref{table_ablation_3} shows that our models maintain computational efficiency compared to the baselines, and the pyramid RoI head only adds little latency per frame.

\begin{table}[]
\setlength{\tabcolsep}{1.9mm}{
\begin{tabular}{|c|c c c|c|}
\toprule
Methods  & R.P. & D.A.R.P. & R.A. & LEVEL\_1 mAP \\
\midrule
PV-RCNN &      &          &      & 70.30           \\
PV-RCNN$^{\star}$ &      &          &      & 74.06         \\
(a)      & $\surd$    &            &            & 75.26           \\
(b)      & $\surd$    & $\surd$    &            & 75.63           \\
(c)      & $\surd$    &            & $\surd$    & 75.77           \\
(d)      & $\surd$    & $\surd$    & $\surd$    & \textbf{76.30}  \\
\bottomrule
\end{tabular}}
\setlength{\belowcaptionskip}{10pt}
\caption{Effects of different components in Pyramid-PV on the Waymo dataset. R.P.: the RoI-grid Pyramid. D.A.R.P.: the Density-Aware Radius Prediction module. R.A.: RoI-grid Attention. $\star$: re-implemented by ourselves with the official code.} \label{table_ablation_1}
\vspace{-2mm}
\end{table}

\begin{table}[]
\setlength{\tabcolsep}{1.7mm}{
\begin{tabular}{|c|c|c|c|}
\toprule
Methods  &  grid size   & $\rho_{w}, \rho_{l}$  & LEVEL\_1 mAP  \\
\midrule
PV-RCNN  & {[}6, 6{]}      & {[}1, 1{]}        & 74.06         \\
(a)      & {[}6,4,4{]}     & {[}1,1,2{]}       & 74.55         \\
(b)      & {[}6,4,4,4{]}   & {[}1,1,2,4{]}     & 74.71         \\
%(c)      & {[}6,4,4,4{]}   & {[}1,1,1.5,2{]}   & 75.13         \\
(c)      & {[}6,4,4,4,1{]} & {[}1,1,1.5,2,4{]} & \textbf{75.26}         \\ 
\bottomrule
\end{tabular}}
\setlength{\belowcaptionskip}{10pt}
\caption{Effects of different RoI pyramids in Pyramid-PV on the Waymo dataset. Each element in [$\cdot$] stands for the respective parameter of a pyramid level.} \label{table_ablation_2}
\vspace{-2mm}
\end{table}

\begin{table}[]
\setlength{\tabcolsep}{6.15mm}{
\begin{tabular}{|l|c|c|c|}
\toprule
Methods  &  Inference speed (Hz)  \\
\midrule
PointRCNN~\cite{shi2019pointrcnn}  &    10.08    \\
\textbf{Pyramid-P (ours)}       &   8.92      \\
\midrule
Part-$A^{2}$ Net~\cite{shi2020points}  & 11.75  \\
\textbf{Pyramid-V (ours)}       &  9.68  \\
\midrule
PV-RCNN~\cite{shi2020pv}  &    9.25      \\
\textbf{Pyramid-PV (ours)}       &  7.86    \\
\bottomrule
\end{tabular}}
\setlength{\belowcaptionskip}{10pt}
\caption{Comparisons on the inference speeds of different detection models on the KITTI dataset.} \label{table_ablation_3}
\vspace{-2mm}
\end{table}

\vspace{-1mm}
\section{Conclusion}
\vspace{-1mm}
We present a general two-stage framework Pyramid R-CNN which can be applied upon diverse backbones. Our framework can handle the sparse and non-uniform distribution problems of point clouds by introducing the pyramid RoI head. For future work, we plan to optimize Pyramid R-CNN for efficient inference.

\clearpage
{\small
\bibliographystyle{ieee_fullname}
\bibliography{egbib}
}
\clearpage

\appendix

\begin{figure}[!t]
\centering
\includegraphics[width=0.45\textwidth]{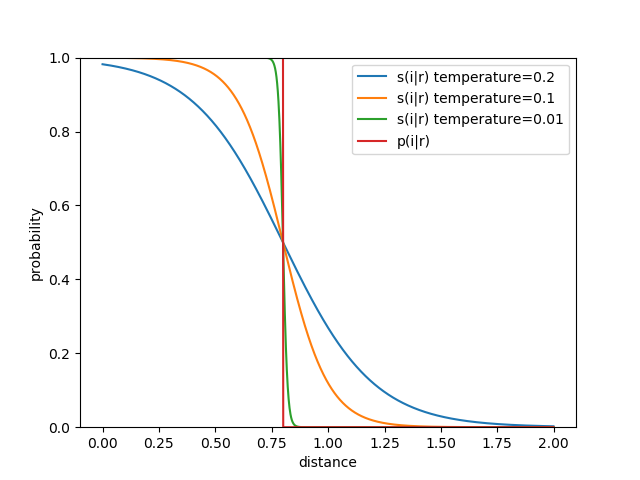}
\caption{Illustration of $p(i|r)$ and $s(i|r)$.}
\label{fig_curve}
\end{figure}

\section{Approximation in Radius Prediction}
In this section, we explain why $s(i|r)$ is used to approximate $p(i|r)$. As is shown in Figure~\ref{fig_curve}, $s(i|r)$ can be viewed as a soft approximation to $p(i|r)$, and the sharpness of the curve $s(i|r)$ is controlled by the temperature $\tau$. When $\tau$ approaches $0$, $s(i|r)$ is more similar to $p(i|r)$. In this paper, we set the initial $\tau$ as $0.02$ for exploration, and gradually decrease $\tau$ to $0.0001$ to obtain a better approximation.

\section{Implementation of the DARP Module}
In this section, we provide the detailed implementation of the Density-Aware Radius Prediction (DARP) module. Inspired by the design of Deformable Convolutions which utilize standard convolutions to predict the deformable offsets, we first use a fixed sphere to aggregate the context embedding and then use this embedding to predict the dynamic radius offset for all grid points in a pyramid level. In particular, for each pyramid level in an \textit{RoI-grid Pyramid}, we use two spheres centered at the RoI with radius $2.4m$ and $4.8m$ for context aggregation. Then the aggregated context embedding is fed into a MLP to predict the dynamic radius offset $\Delta r$. A predefined radius $r$ added by the dynamic offset $\Delta r$, \ie $r + \Delta r$, is utilized to obtain the coefficient $s(i, r+\Delta r)$ in Eq.10, and with $s(i, r+\Delta r)$, the \textit{Points of Interest} within $r + \Delta r + 5\tau$ are selected as $\Omega(r + \Delta r + 5\tau)$ for the computation of \textit{RoI-grid Attention} in Eq.13, where $\tau$ is the temperature. The predefined $r$ in this paper is set to $0.8$, $1.6$, $2.4$, $3.2$, $6.4$ for the respective pyramid level. We note that all grid points in a pyramid level share the same $r + \Delta r$, and the prediction of $\Delta r$ adds little computational overhead. It is worth noting that we can easily extend this idea to the settings where each grid point has its individual predicted radius, or we can additionally predict centers for the predicted spheres. 

\begin{figure}[!t]
\centering
\includegraphics[width=0.45\textwidth]{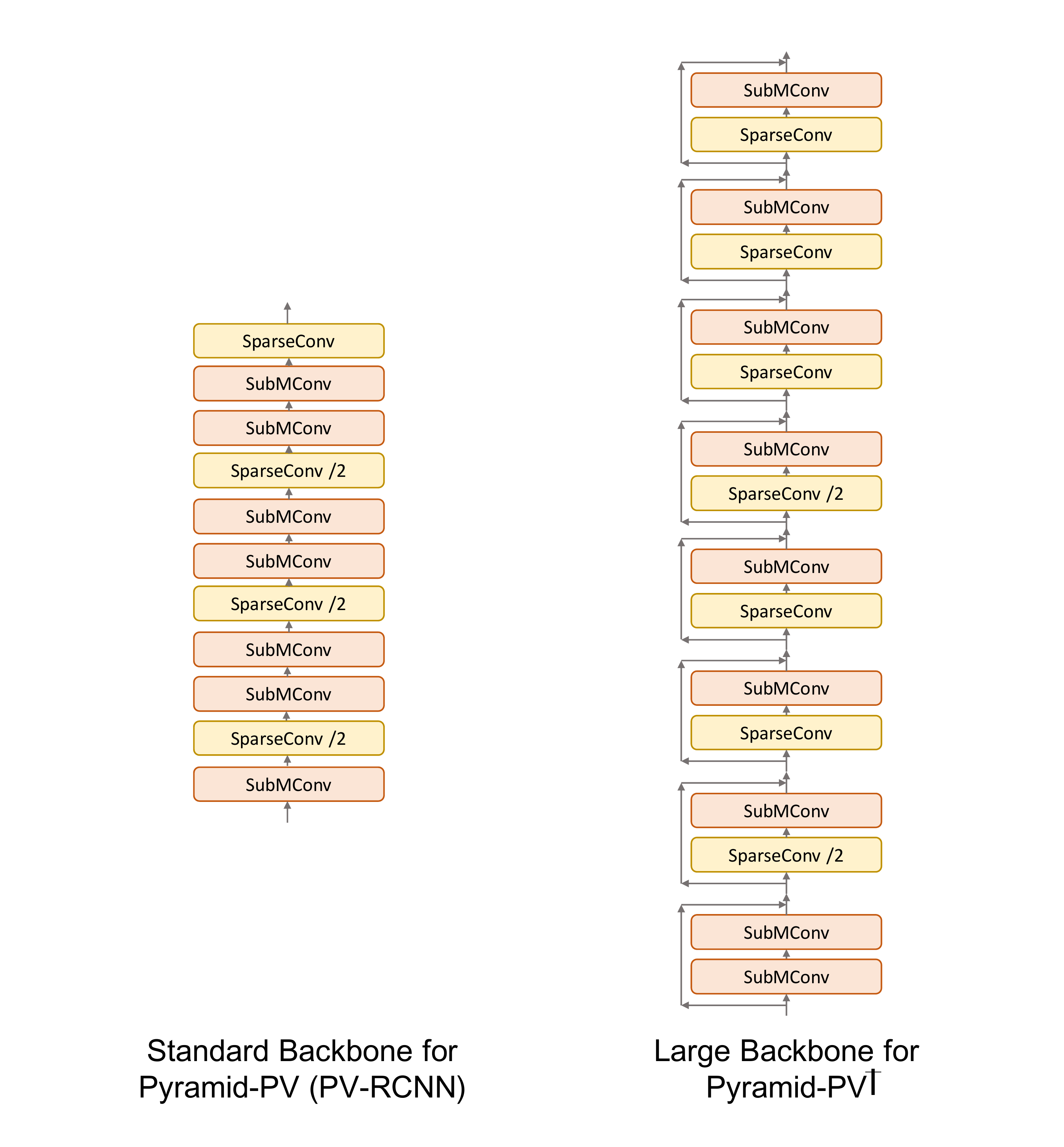}
\caption{3D Backbone of Pyramid-PV and Pyramid-PV$^{\dag}$.}
\label{fig_backbone}
\vspace{-3mm}
\end{figure}

\section{Backbones of Pyramid R-CNN}
In this section, we provide additional information for some backbones of Pyramid R-CNN. We note that other backbones that are not mentioned are directly referred from the official source-code repositories. \textit{Pyramid RoI head} is kept the same upon all the backbones in this paper.

\textbf{Pyramid-P.} We re-implement the backbone of PointRCNN on the Waymo Open dataset. Different from the original version on the KITTI dataset, we set the number of input sampled point clouds to $40k$, and the number of downsampled points is set to $18024$, $2048$, $512$, $128$ for the respective layer. We note that this modification enlarges the number of kept points, since the number of input point clouds is larger compared with those on the KITTI dataset. \textit{Pyramid-P} and our re-implemented PointRCNN share the same backbone configurations on the Waymo Open dataset.    

\textbf{Pyramid-PV$^{\dag}$.} In \textit{Pyramid-PV$^{\dag}$} we implement a larger backbone with the input voxel size as $[0.2m, 0.2m, 0.2m]$. The backbones of \textit{Pyramid-PV$^{\dag}$} and vanilla \textit{Pyramid-PV} (PV-RCNN) are shown in Figure~\ref{fig_backbone}. Our Pyramid R-CNN is compatible with a small backbone for a fair comparison with baseline methods, or a large backbone to further enhance the detection performance.

\section{Qualitative Results}
\begin{figure*}[!t] \centering   
\subfigure[] {
 \label{viz_1_1}     
\includegraphics[width=0.4\textwidth]{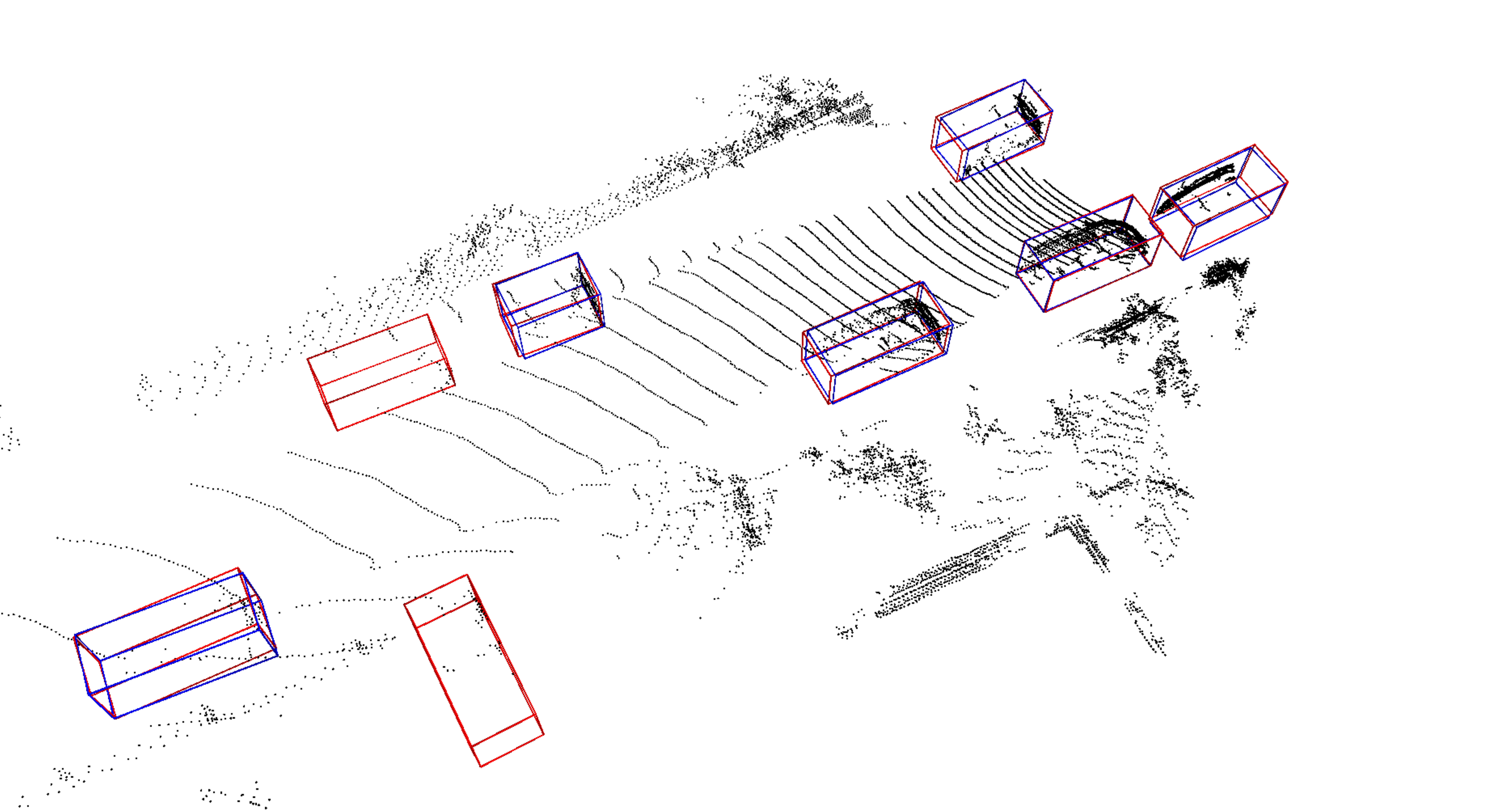}  
}    
\subfigure[] { 
\label{viz_1_2}     
\includegraphics[width=0.4\textwidth]{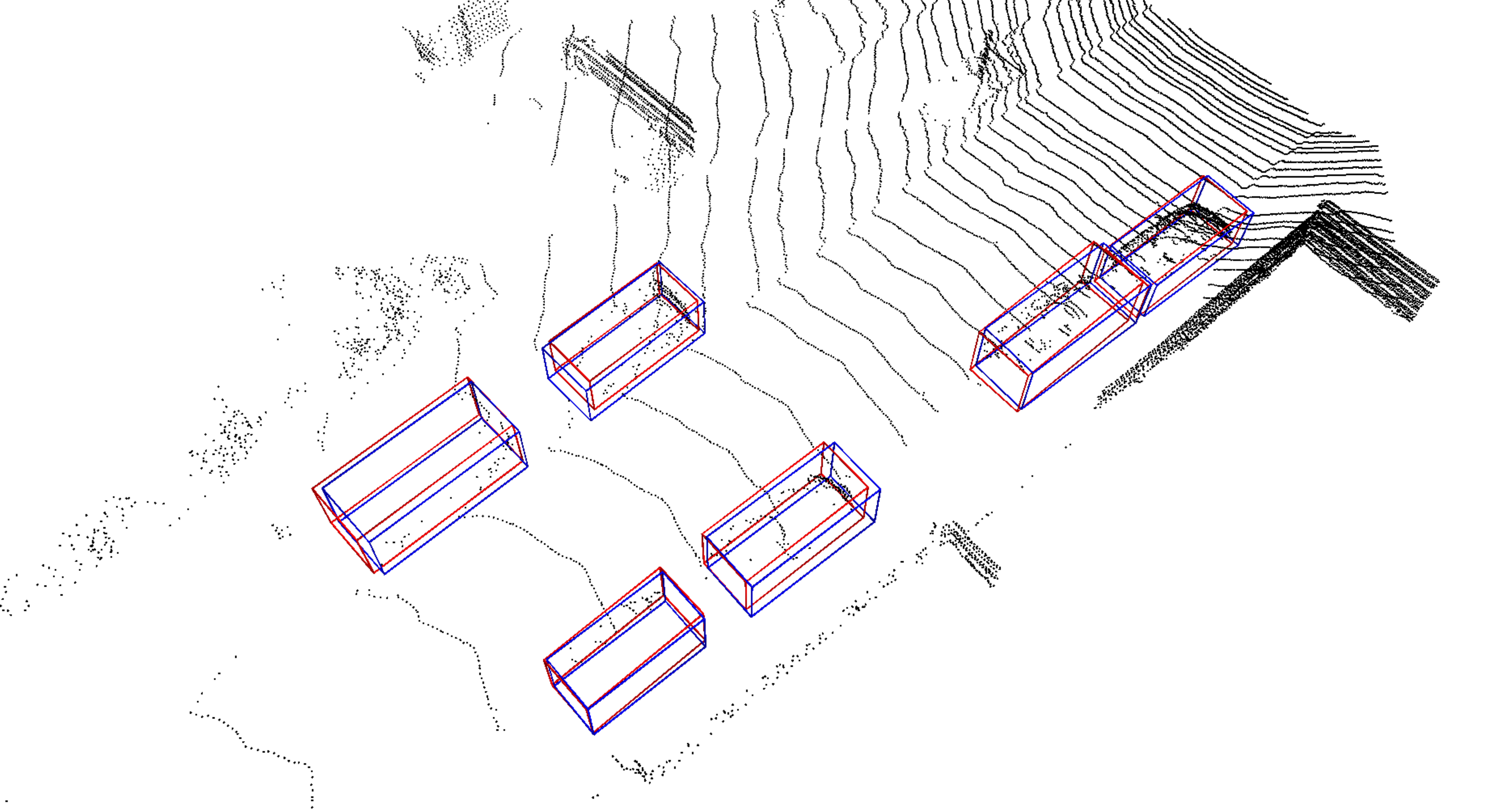}     
}
\subfigure[] { 
\label{viz_1_3}     
\includegraphics[width=0.4\textwidth]{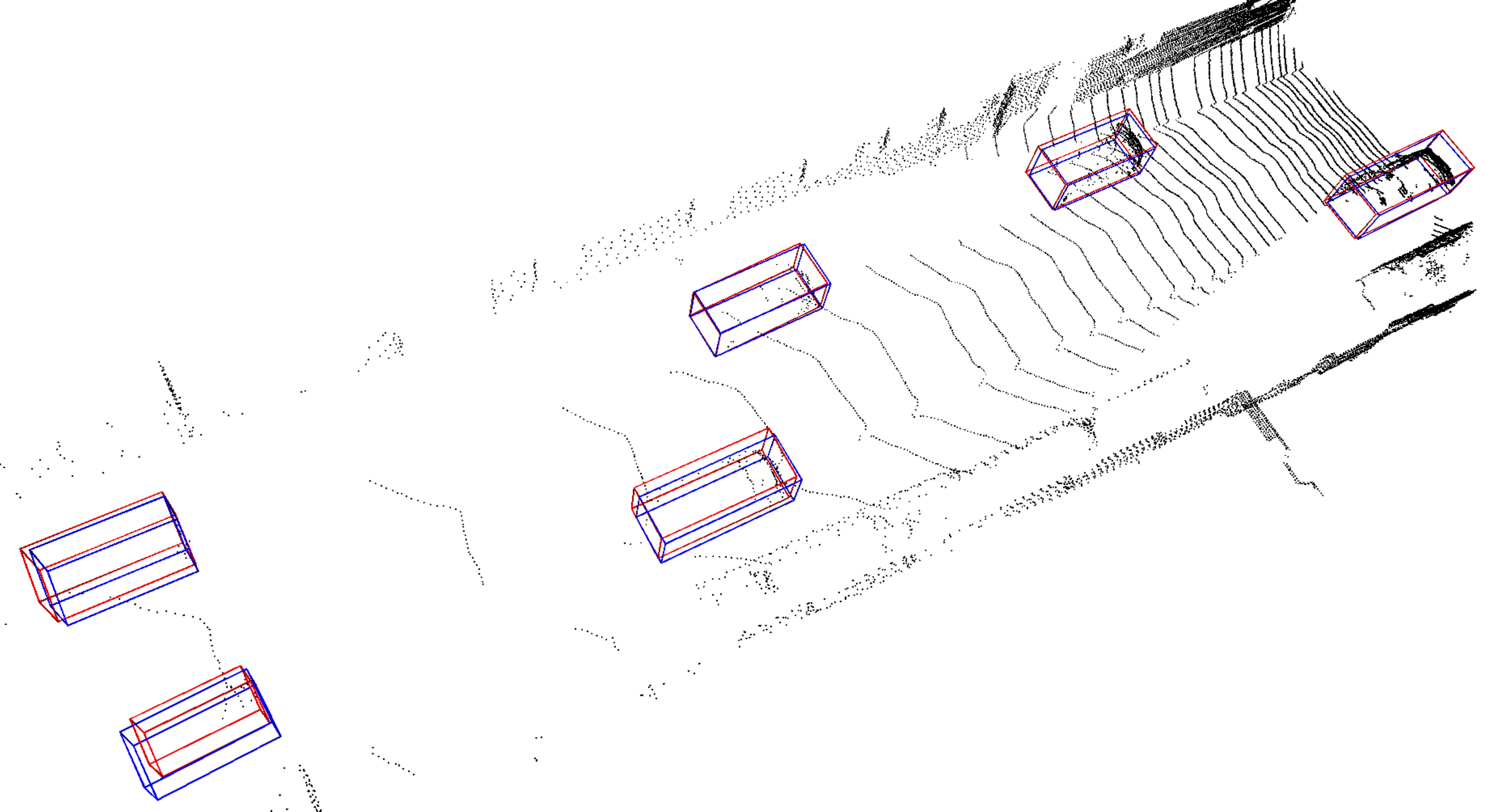}     
}
\caption{Visualization of detection results on the KITTI dataset. Blue boxes are the ground truth boxes, and red boxes are the boxes predicted by \textit{Pyramid-PV}.}     
\label{fig_viz_1} 
\end{figure*}

\begin{figure*}[!t] \centering   
\subfigure[] {
 \label{viz_2_1}     
\includegraphics[width=0.4\textwidth]{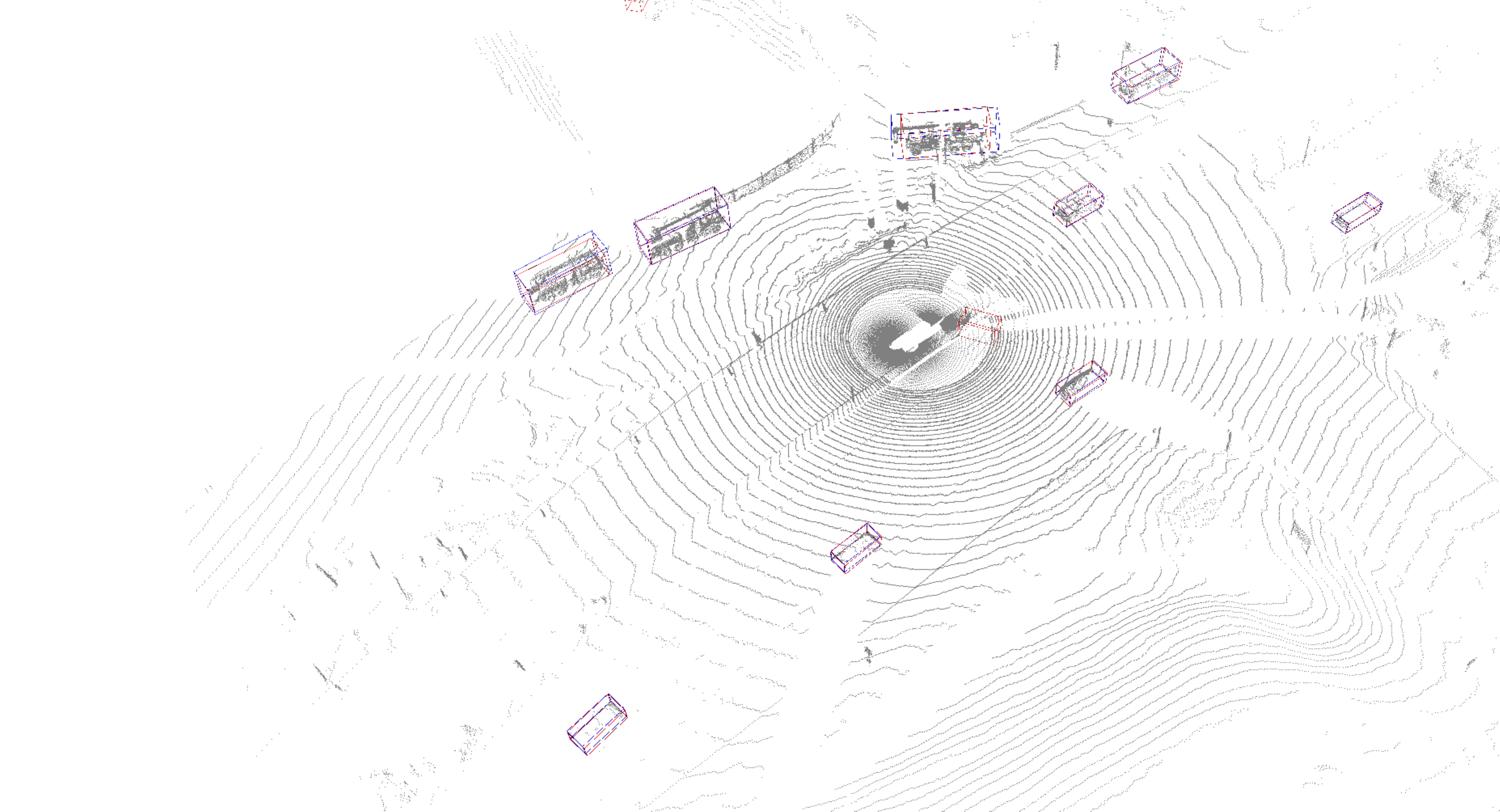}  
}    
\subfigure[] { 
\label{viz_2_2}     
\includegraphics[width=0.4\textwidth]{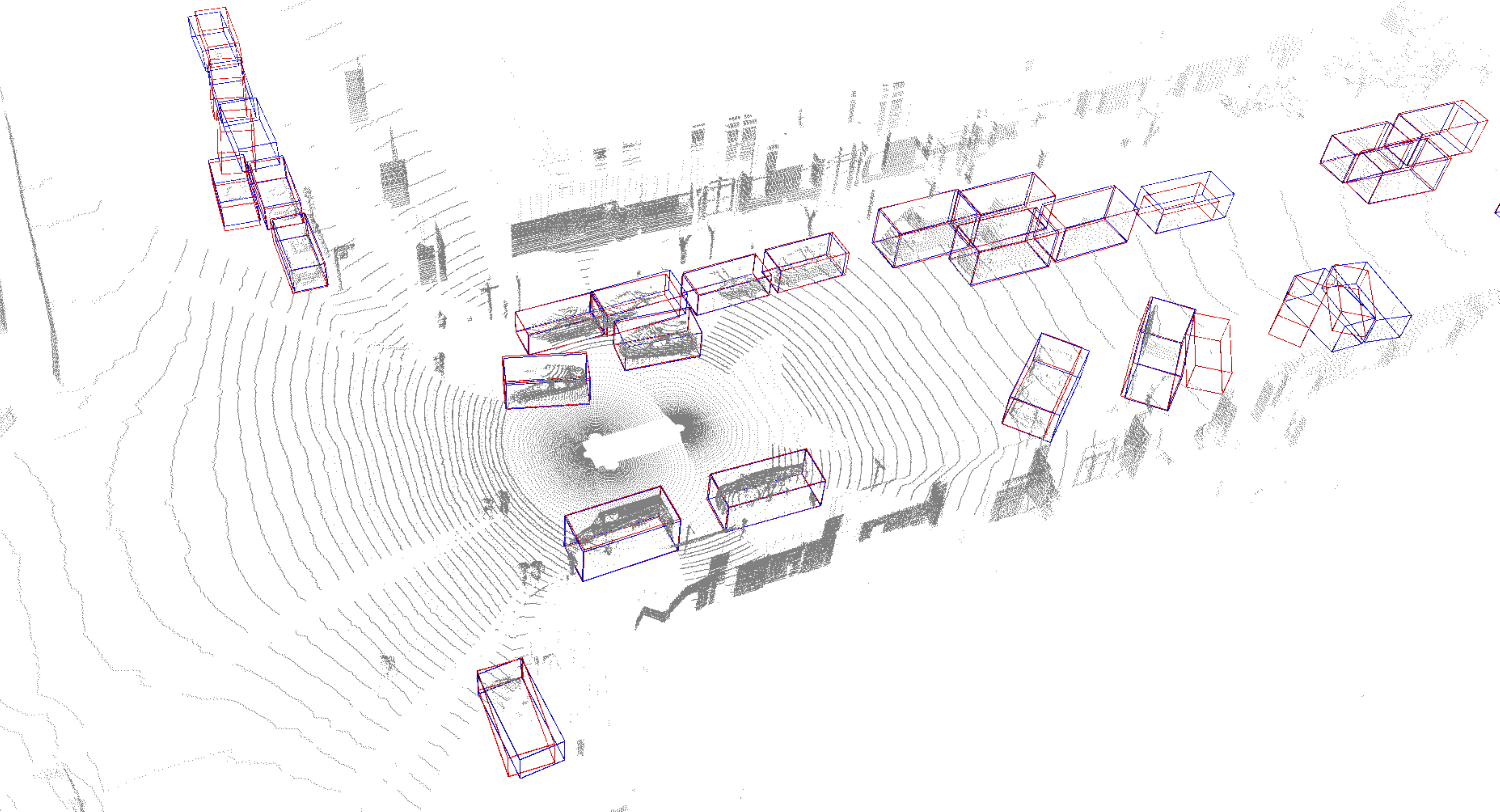}     
}
\subfigure[] { 
\label{viz_2_3}     
\includegraphics[width=0.4\textwidth]{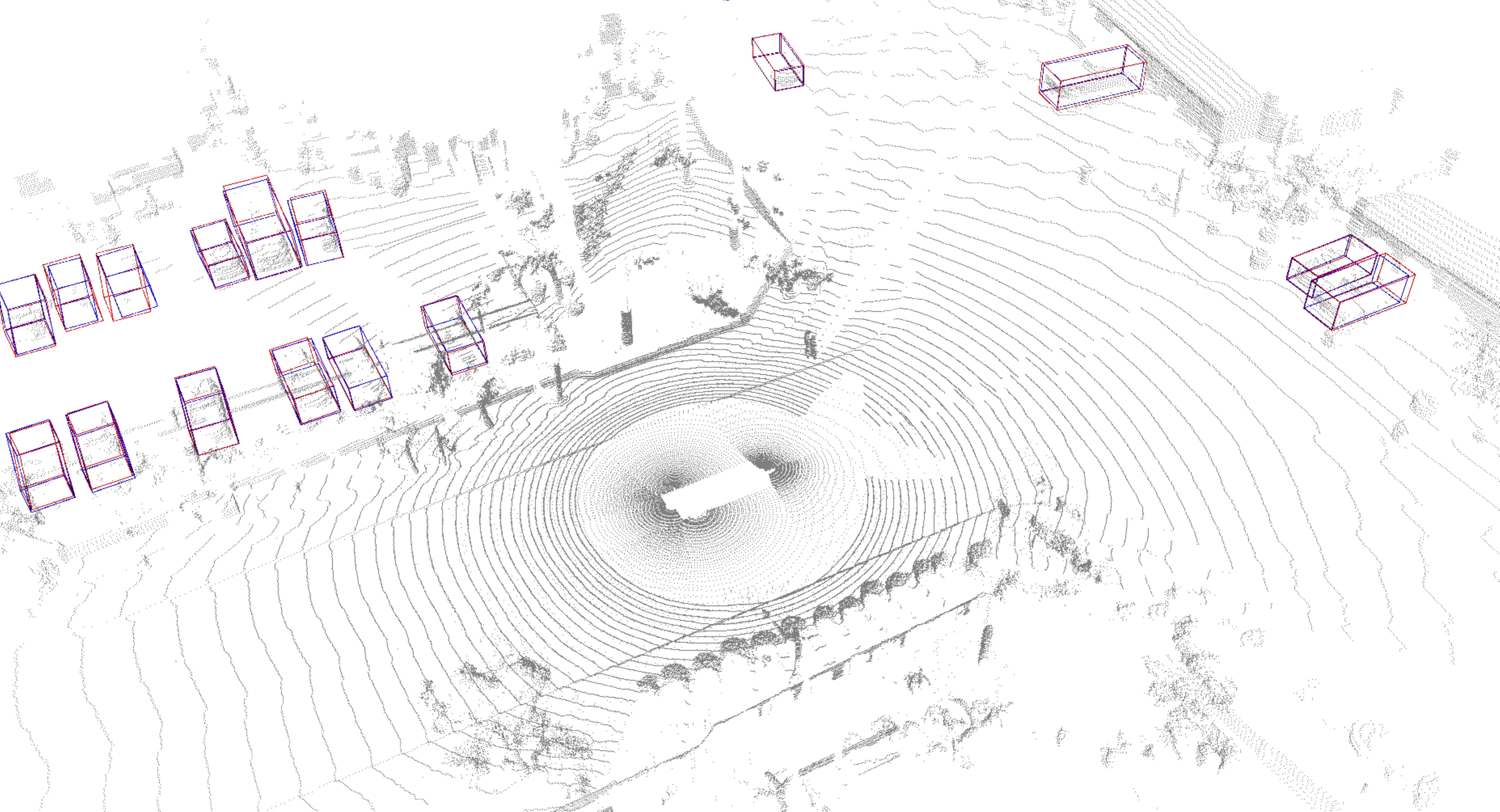}     
}
\caption{Visualization of detection results on the Waymo Open dataset. Blue boxes are the ground truth boxes, and red boxes are the boxes predicted by \textit{Pyramid-PV}.}     
\label{fig_viz_2} 
\end{figure*}

In this section, we provide the qualitative results on the KITTI dataset in Figure~\ref{fig_viz_1}, and the Waymo Open dataset in Figure~\ref{fig_viz_2}. The figures show that our proposed Pyramid R-CNN can accurately detect 3D objects which are far away and have only a few points.  

\end{document}